\documentclass[journal]{IEEEtran}

\usepackage[export]{adjustbox}
\usepackage{amssymb}
\usepackage{bm}
\usepackage{booktabs}
\usepackage{cite}
\usepackage{enumitem}
\usepackage{glossaries}
\usepackage[bookmarks=true]{hyperref}
\usepackage{siunitx}
\usepackage{subcaption}
\usepackage[normalem]{ulem}

\graphicspath{{figures/}}

\hypersetup{hidelinks}

\DeclareMathOperator*{\argmax}{\operatornamewithlimits{\arg\!\max}}

\hypersetup{
    pdfauthor   = {Henrique Ferrolho},%
    pdftitle    = {RoLoMa: Robust Loco-Manipulation for Quadruped Robots with Arms},%
    pdfkeywords = {Robust, Loco-Manipulation, Optimization, Polytopes},%
    pdfsubject  = {Robotics},%
}

\begin{document}

\newacronym{CoM}{CoM}{center of mass}
\newacronym{DoF}{DoF}{degrees of freedom}
\newacronym{GIWC}{GIWC}{Gravito-Inertial Wrench Cone}
\newacronym{LQR}{LQR}{linear-quadratic regulator}
\newacronym{MPC}{MPC}{model predictive control}
\newacronym{MRP}{MRP}{modified Rodrigues parameters}
\newacronym{NLP}{NLP}{nonlinear programming}
\newacronym{RBD.jl}{RBD.jl}{RigidBodyDynamics.jl}
\newacronym{RL}{RL}{reinforcement learning}
\newacronym{RMSE}{RMSE}{root-mean-square error}
\newacronym{SUF}{SUF}{smallest unrejectable force}
\newacronym{TO}{TO}{trajectory optimization}
\newacronym{wrt}{w.r.t.}{with respect to}

\title{RoLoMa: Robust Loco-Manipulation \\ for Quadruped Robots with Arms}

\author{
    Henrique Ferrolho${}^{1}$, Vladimir Ivan${}^{2}$, Wolfgang Merkt${}^{3}$, Ioannis Havoutis${}^{3}$, Sethu Vijayakumar${}^{4}$%
    \thanks{${}^{1}$Ocado Technology, Welwyn Garden City, UK.}%
    \thanks{${}^{2}$Touchlab Limited, Edinburgh, UK.}%
    \thanks{${}^{3}$Oxford Robotics Institute, University of Oxford, UK.}%
    \thanks{${}^{4}$School of Informatics, University of Edinburgh, UK.}%
    \thanks{This research was done when Henrique Ferrolho and Vladimir Ivan were with the School of Informatics, University of Edinburgh, UK.}%
    \thanks{\textit{Email address:} {\href{mailto:henrique.ferrolho@gmail.com}{\nolinkurl{henrique.ferrolho@gmail.com}}}}%
}

\maketitle

\begin{abstract}
    Deployment of robotic systems in the real world requires a certain level of robustness in order to deal with uncertainty factors, such as mismatches in the dynamics model, noise in sensor readings, and communication delays.
    Some approaches tackle these issues \emph{reactively} at the control stage.
    However, regardless of the controller, online motion execution can only be as robust as the system capabilities allow at any given state.
    This is why it is important to have good motion plans to begin with, where robustness is considered \emph{proactively}.  %
    To this end, we propose a metric (derived from first principles) for representing robustness against external disturbances.
    We then use this metric within our trajectory optimization framework for solving complex loco-manipulation tasks.
    Through our experiments, we show that trajectories generated using our approach can resist a greater range of forces originating from any possible direction.
    By using our method, we can compute trajectories that solve tasks as effectively as before, with the added benefit of being able to counteract stronger disturbances in worst-case scenarios.
\end{abstract}


\IEEEpeerreviewmaketitle

\begin{figure}[t]
    \captionsetup{font=small}
    \centering
    \begin{subfigure}[t]{\linewidth}
        \includegraphics[width=0.498\linewidth]{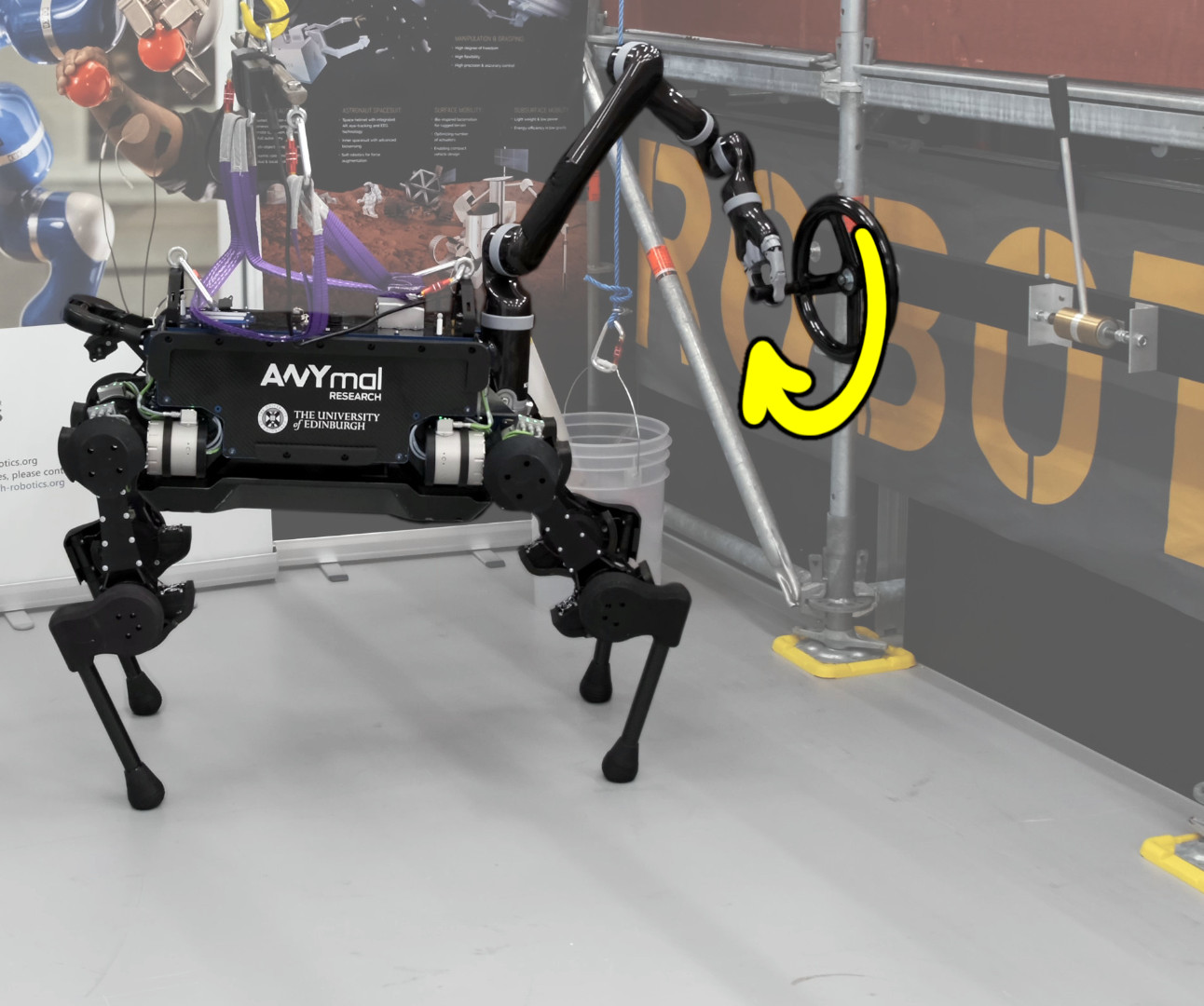}\hfill%
        \includegraphics[width=0.498\linewidth]{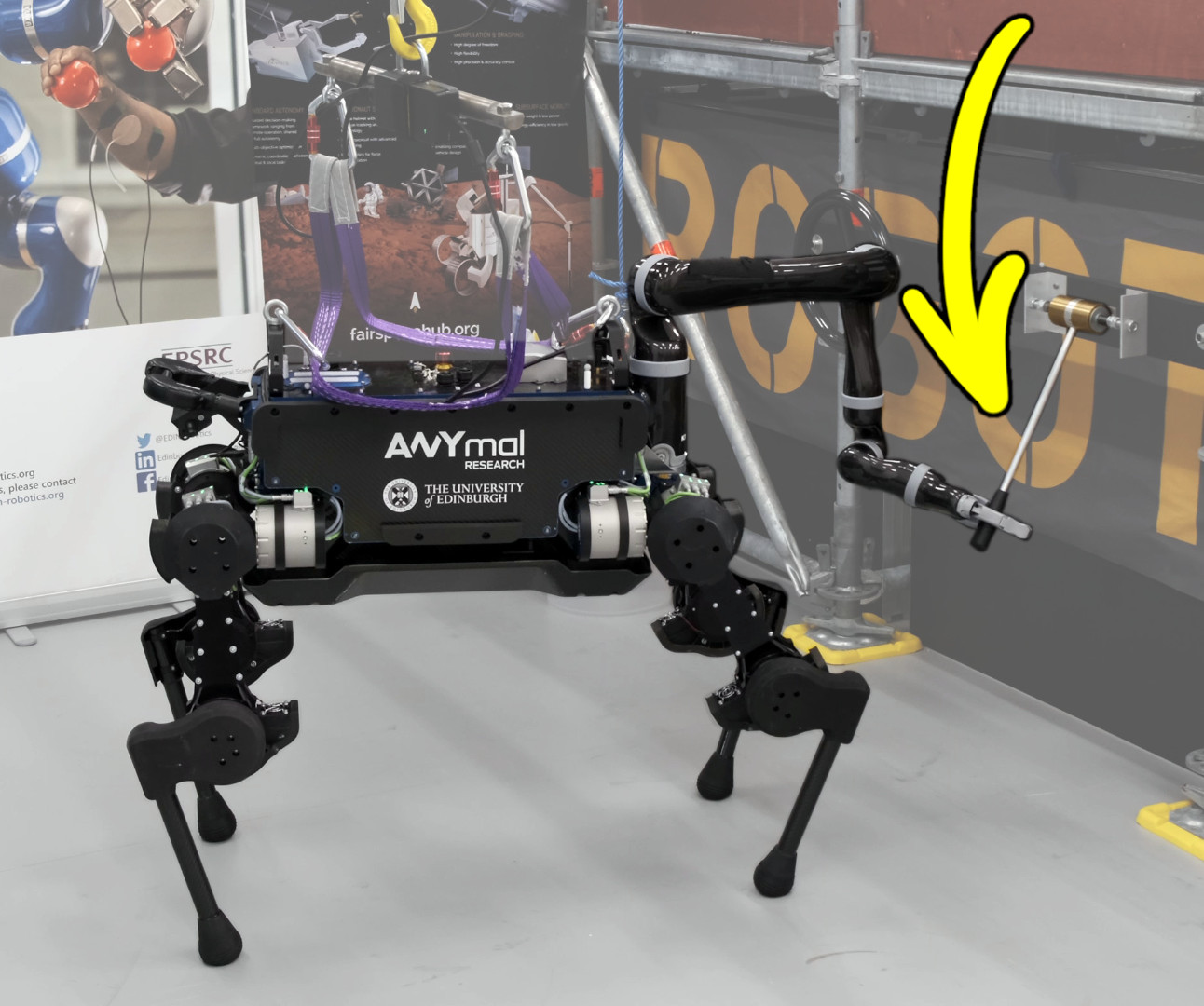}\hfill%
    \end{subfigure}\hfill%
    \vspace{0.004\linewidth}
    \begin{subfigure}[t]{\linewidth}
        \includegraphics[width=0.498\linewidth]{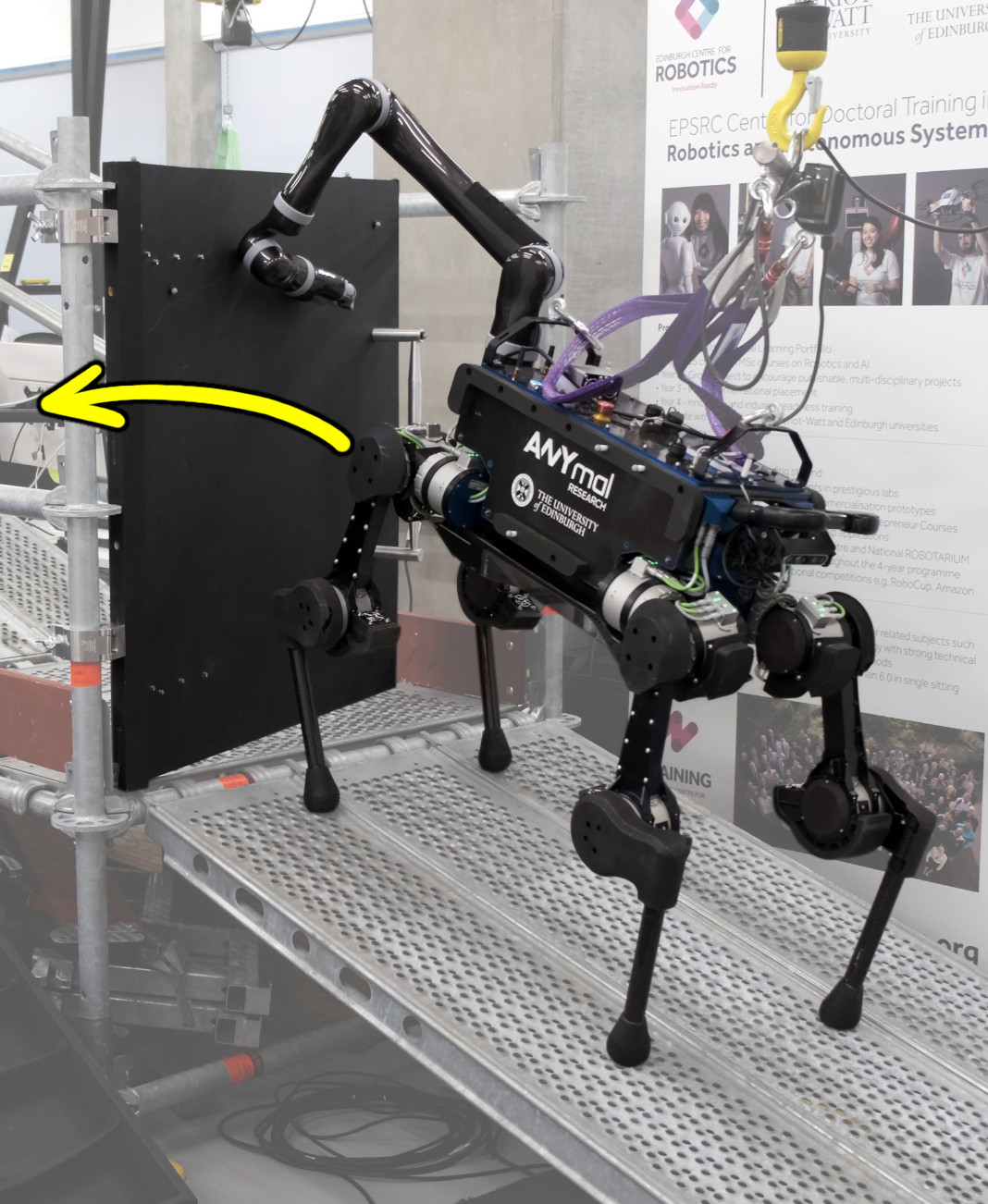}\hfill%
        \includegraphics[width=0.498\linewidth]{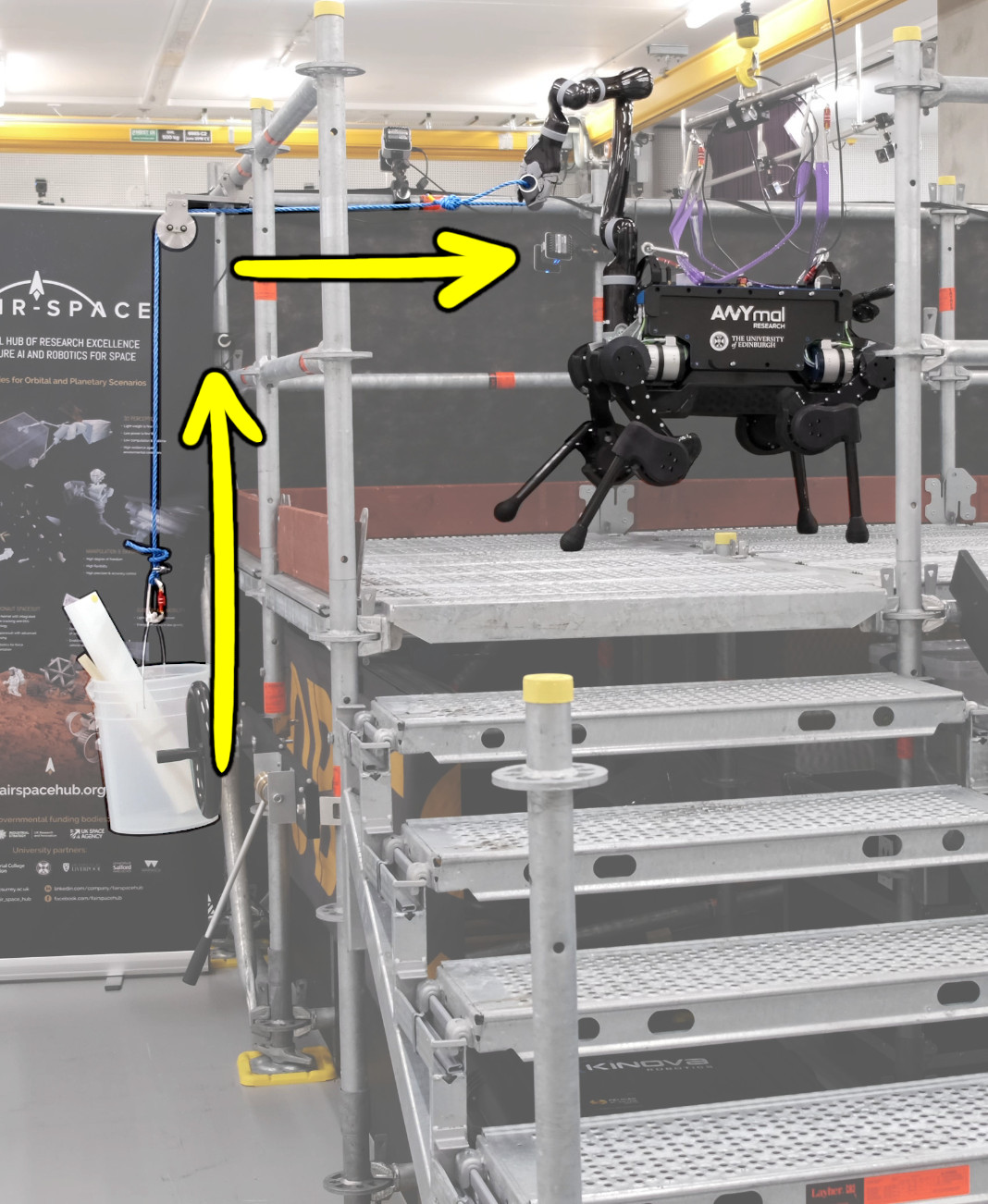}\hfill%
    \end{subfigure}
    \caption{
        Snapshots of our robot solving real-world tasks in an industrial setting: (i) turning a hand wheel, (ii) pulling a lever, (iii) opening a gate whilst standing on a ramp, and (iv) lifting a bucket by pulling a rope.
        The robot and the objects being manipulated have been highlighted for clarity.
        The overlaid yellow arrows indicate motion.
        Video footage: \texttt{\small\url{https://youtu.be/3qXNHVCagL8}}.
    }\label{figure:cover}
\end{figure}

\section{Introduction}
\label{sec:introduction}

In this paper, we tackle the problem of robust loco-manipulation for quadruped robots equipped with robot arms, such as the one shown in \autoref{figure:cover}.
Here, the challenge is not only to generate whole-body trajectories for solving complex tasks requiring simultaneous locomotion and manipulation (commonly referred to as \emph{loco-manipulation}), but also to optimize the robustness of such trajectories against unknown external disturbances.
This is an important problem for two reasons:
first, loco-manipulation allows us to extend the workspace of an otherwise-fixed-base-manipulator through the mobility of a mobile base, such as a legged robot;
and second, increasing the robustness of the overall motion against disturbances leads to legged systems that are more reliable, and that can therefore be deployed in the real world with greater confidence.

Enabling loco-manipulation is a very challenging problem because it involves repeatedly breaking and making contacts between the feet and the environment in order to move around, while maintaining balance and avoiding kinematic/actuation limits.
Moreover, robust motion planning is a complex subject on its own, as it requires the derivation of good metrics that are able to quantify how robust trajectories are.
Consequently, combining loco-manipulation and robustness is not trivial, as it brings together the challenges from both problems.

Most of the previous research done on loco-manipulation, \cite{murphy2012high,zimmermann2021go,ma2022combining}, has tackled the problem by splitting the arm from the base, planning the manipulation separately from the locomotion, and then considering arm movement as a disturbance that the base should compensate for.
Furthermore, most of the existing research \cite{prete2016robustness,xin2018model,sleiman2021unified,ma2022combining} has focused on reactive robustness at the control stage, rather than taking into account robustness proactively during planning.

The key components of our approach are
(i) a trajectory optimization framework which is able to solve complex real-world tasks and which takes into account the full system dynamics, and
(ii) a robustness metric derived from first principles.
This allows us to calculate the largest force magnitude that the robot can counteract from any given direction, while considering ground-feet contact stability and the actuation limits of the system.
Our results show that, given a contact sequence, our framework is able to plan whole-body trajectories that are significantly more robust to external disturbances compared to other approaches.

\subsection{Statement of Contributions}
This paper is a direct follow-up of our previous work.
In \cite{ferrolho2020optimizing}, the framework we proposed was able to optimize whole-body trajectories for standing balance behaviors only, i.e., not \emph{actual} loco-manipulation.
In contrast, this paper proposes an improved formulation which is able to optimize trajectories that involve contact changes---albeit the sequence and timings of those contacts must be prescribed. %
We are also able to better enforce the nonlinear dynamics of rigid-body systems, as we have incorporated our findings from \cite{ferrolho2021inverse}.
Furthermore, we show that our formulation can handle cases where contact positions are not enforced explicitly, which actually allows it to further maximize robustness through the adoption of more suitable feet contact positions.
Finally, the significant amount of systems integration work that we have done in comparison to \cite{ferrolho2020optimizing} allowed us to deploy the robot in a realistic scenario mimicking an industrial offshore platform, where we showed the robot operating uninterrupted and repeatedly solving the complex sequence of tasks highlighted in \autoref{figure:cover}.
This has resulted in a robust loco-manipulation system which is capable of online motion planning for deployment in realistic scenarios.

\section{Related Work}

We now summarize previous research related to motion planning for quadruped robots equipped with arms, as well as existing research on the topic of robustness.

\subsection{Planning and Control for Quadrupeds with Arms}

Murphy \textit{et al.} \cite{murphy2012high} were one of the first to investigate the use of a legged robot base to improve the capabilities of a robotic arm.
They used \gls{TO} with a simplified dynamics model to generate open-loop behaviors for Boston Dynamics' robot BigDog.  %
As a result of the coordinated motion between the robot arm and the robot base, they were able to increase the performance of lifting and throwing tasks.
In their hardware experiments, they showed the robot dynamically tossing cinder blocks as heavy as \SI{16.5}{\kilo\gram} and as far as \SI{4.2}{\metre}.
However, they only considered standing balance behaviors (which do not change support contacts/move the feet).
In contrast, we consider the full dynamics model of the robot during \gls{TO}, and we also consider behaviors where the robot's feet can make and break contact with the environment.

A few years later, Zimmermann \textit{et al.} \cite{zimmermann2021go} equipped Boston Dynamics' flagship quadruped robot Spot with a Kinova arm in order to perform dynamic grasping maneuvers.
Direct control of Spot's actuated joints is not possible because of restricted access to its low-level controller.
As Spot's whole-body controller for locomotion compensates for the wrench induced by the manipulator, the tracking and executed motion can differ from planned motion, resulting in poor performance.
To achieve precise loco-manipulation, Zimmermann \textit{et al.} treated Spot's overall behavior as a black box, and built a simplified model of the combined platform from experimental data.
While their approach successfully grasped the target most of the times, it failed when the estimated position of the ball was inaccurate or when the robot started executing the planned trajectories from slight offset poses.
There were also cases when the robot failed to grasp the target due to Spot's complex internal behavior not being fully captured by their simplified model.
For example, when the disturbance induced by the arm to the base was sufficiently large, it could cause a delay that brought the individual trajectory components out of sync.
In contrast, our approach does not suffer from this drawback because the robot we use in our experiments grants us full control over its joints, and because we optimize trajectories for the robot considering the dynamics of its whole body.

Ma \textit{et al.} \cite{ma2022combining} combined manipulation using \gls{MPC} with a locomotion policy obtained from \gls{RL}.
First, they modeled the wrenches (arising from the motion of the arm) applied to the base of the robot as external disturbances that can be predicted.
Then, they trained the base control policy to counteract those disturbances while
(i) trying to keep a horizontal base orientation and
(ii) tracking velocity commands from the \gls{MPC} controller of the arm.
In other words, their base policy uses wrench predictions from the arm's motion to compensate for the disturbances applied to the base of the robot.

All of the previous work mentioned thus far \cite{murphy2012high,zimmermann2021go,ma2022combining} have one thing in common:
they all see the arm as a disturbance to be compensated for.
In contrast, Bellicoso \textit{et al.} \cite{bellicoso2019alma} approached the problem differently, and took into account the dynamics of the whole system.
The authors used a whole-body controller based on inverse dynamics, re-planned locomotion continuously in a receding-horizon fashion, and explicitly provided end-effector forces for the controller to track.
They equipped ANYbotics' quadruped robot ANYmal B with a Kinova arm---a combination which results in a fully torque-controlled mobile manipulator, and one which users have full control over.  %
Bellicoso \textit{et al.} demonstrated that the resulting system is able to perform dynamic locomotion while executing manipulation tasks, such as opening doors, delivering payloads, and human-robot collaboration.

In \cite{murphy2012high}, the manipulation task is planned offline and separately from the locomotion planner;
and in \cite{bellicoso2019alma}, the task for opening the door uses a controller that tracks gripper forces which need to be explicitly specified.
Sleiman \textit{et al.} \cite{sleiman2021unified} tackled both these weaknesses when they proposed a unified \gls{MPC} framework for whole-body loco-manipulation.
Their approach augments the dynamics of the object being manipulated to the centroidal dynamics and full kinematics of the robot.
This allows the solver to exploit the base-limb coupling and, e.g., to use the arm as a balancing ``tail''.
Despite using an \gls{MPC} approach, their planner is not adaptive with respect to the dynamic properties of the objects being manipulated.

In our previous work \cite{ferrolho2020optimizing}, we proposed a motion planning framework for legged robots equipped with manipulators.
In that work---and in this paper---we used the same robot arm and quadruped shown in \cite{bellicoso2019alma}, with a difference only in the number of fingers on the gripper.
Our motion planning approach was similar to \cite{sleiman2021unified} in the sense that we formulated the planning problem for the whole body of the robot in a unified manner, i.e., for the quadrupedal base and the robot manipulator simultaneously.
There are a couple of differences in the problem formulation between our previous work \cite{ferrolho2020optimizing} and \cite{bellicoso2019alma,sleiman2021unified}; e.g., we use a full model of the robot's articulated rigid-body dynamics\footnote{Henceforth, we shall refer to this as the `full dynamics model' of the robot.} instead of a simplified version, which allows us to plan trajectories more faithfully to the real hardware.
However, the most important difference and our main contribution is that we focus on planning motions that are not only physically feasible, but that also maximize robustness against unknown external disturbances.
This aspect is something that none of the previous work mentioned \cite{murphy2012high,zimmermann2021go,ma2022combining,bellicoso2019alma,sleiman2021unified} have considered.

\subsection{Motion Robustness Against Disturbances}

Del Prete \textit{et al.} \cite{prete2016robustness} proposed a solution to improve the robustness to joint-torque tracking errors at the control stage.
The authors modeled deterministic and stochastic uncertainties in joint torques within their control framework optimization.
In our case, we maximize the upper-bound force magnitude the system can withstand from any possible direction, and we do this during the planning stage.
Xin \textit{et al.} \cite{xin2018model} proposed a hierarchical controller in which external forces are estimated directly.
Their goal was to minimize actuator torques while enforcing constraints for the contact forces.
However, in contrast to our work, they did not explicitly enforce actuator limits; and since their main focus is on control, they do not have a planner for computing elaborate whole-body behaviors.

The robot experiments shown in \cite{ma2022combining} and in \cite{sleiman2021unified} demonstrate some capability of resistance against external forces; but this robustness is \emph{reactive}, in the sense that it is either the learned locomotion policy or the \gls{MPC} that compensate for the disturbances online in a reactive fashion.
In contrast, research on \emph{proactive} robustness \cite{caron2015leveraging,orsolino2018application,ferrolho2021residual} attempts to take uncertainty into account at the planning stage, i.e., ahead (or just in time) of online execution.
A proactive approach allows planning frameworks to increase the system's ability of counteracting disturbances by exploiting kinematic redundancy.
Some examples of human kinematics pre-shaping during dynamic tasks are: leaning body backwards while pulling on a rope during a game of tug of war, leaning body forwards and stretching arms while pushing on a broken down car, and a bent-knees stance in preparation for a skateboard trick.
Similarly, in robotics, we can increase robustness proactively with offline planning, by e.g. compiling a library of optimal kinematic stances and optimal trajectories, which can then be looked up during online execution.

In \cite{caron2015leveraging}, Caron \textit{et al.} proposed a feasible region (a \emph{polytope}), called \gls{GIWC}, which can be used as a general stability criterion.
This representation is very efficient for testing the robust static equilibrium of a legged robot, but it neglects the system's actuation limits.
Subsequently, Orsolino \textit{et al.} \cite{orsolino2018application} proposed to extend the properties of the \gls{GIWC} by incorporating the torque limits of the system.
They demonstrated how the resulting polytopes can be used to e.g. optimize the \gls{CoM} trajectory in the $xy$-plane for the base-transfer motion of quadrupeds.
Orsolino \textit{et al.} formulated a reduced version of the problem, but even then the technique used to compute polytopes was prohibitively expensive.
As a workaround, they computed the polytope only once for the first point of the trajectory, and used that as an approximation for the rest of the motion.
We have followed this line of research in our previous work \cite{ferrolho2021residual}, where we proposed a force polytope representation, called \emph{residual force polytope}, which considers not only the torque limits but also the dynamics of the system during trajectory execution.
The polytope is computed from the forces and torques remaining after accounting for Coriolis, centrifugal, and gravity terms, as well as from the nominal feed-forward torques of the motion.

The polytope calculations in \cite{orsolino2018application} and \cite{ferrolho2021residual} require significant computation time and, in general, deriving explicit descriptions of a projected polytope is NP-hard \cite{tiwary2008hardness}.
Zhen \textit{et al.} \cite{zhen2018computing} formulated a computationally-tractable approach for finding maximally-sized convex bodies inscribed in projected polytopes.
Later, Wolfslag \textit{et al.} \cite{wolfslag2020optimisation} adapted that approach for computing the robustness of static robot configurations.
Their work was an improvement over exact computations; however, it was still too complex for being considered in trajectory optimization.
In our previous work \cite{ferrolho2020optimizing}, we adapted the technique from \cite{zhen2018computing} to reformulate the problem of computing the \gls{SUF}, which allowed us to formulate bilevel trajectory optimization problems for maximizing the robustness of the generated trajectories.
In this paper, our work extends those ideas further to motion with contact changes.

\section{Robust Trajectory Optimization}
\label{sec:robust_trajectory_optimization}

\subsection{Robot Model Formulation}
We formulate the model of a legged robot in the same way as in our previous work \cite{ferrolho2020optimizing}, i.e., as a free-floating base $B$ to which the limbs are attached to.
For example, the robot we used for our experiments (seen in \autoref{figure:cover}), has four legs and one arm attached to its base; each leg has three motors and the arm has six.\footnote{
    Please note that our formulation is not tied to the specific robot shown in \autoref{figure:cover}.
    In fact, it is general enough such that it can be applied to any legged robot, biped or quadruped, with or without arms.
}
We describe the motion of the system \gls{wrt} a fixed inertial frame $I$.
We represent the position of the free-floating base \gls{wrt} the inertial frame, and expressed in the inertial frame, as ${}_I\bm{r}_{IB} \in \mathbb{R}^3$.
As for the orientation of the base, we represent it using \gls{MRP}~\cite{gormley1945stereographic,terzakis2018modified} as $\bm{\psi}_{IB} \in \overline{\mathbb{R}}{}^3$.
The joint angles describing the configuration of the 6-\gls{DoF} arm and the four 3-\gls{DoF} legs are stacked in a vector $\bm{q}_j \in \mathbb{R}^{n_j}$, where $n_j = 18$.
Finally, we write the generalized coordinates vector $\bm{q}$ and the generalized velocities vector $\bm{v}$ as
\begin{equation}
    \bm{q} = \begin{bmatrix} {}_I\bm{r}_{IB} \\ \bm{\psi}_{IB} \\ \bm{q}_j \end{bmatrix} \in \mathbb{R}^3 \times \overline{\mathbb{R}}^3 \times \mathbb{R}^{n_j}, \quad
    \bm{v} = \begin{bmatrix} \bm{\nu}_B \\ \bm{\dot{q}}_j \end{bmatrix} \in \mathbb{R}^{n_v},
\end{equation}
where the twist $\bm{\nu}_B = \left [ {}_I\bm{v}_{B} \quad {}_B\bm{\omega}_{IB} \right ]^\top \in \mathbb{R}^6$ encodes the linear and angular velocities of the base $B$ w.r.t. the inertial frame expressed in the $I$ and $B$ frames, and $n_v = 6 + n_j$.

The equations of motion of a floating-base rigid-body system that interacts with the environment are written as
\begin{equation}
    \bm{M}(\bm{q})\bm{\dot{v}} + \bm{h}(\bm{q}, \bm{v}) = \bm{S}^\top \bm{\tau} + \bm{J}_s^\top(\bm{q}) \bm{\lambda} + \bm{J}_e^\top(\bm{q}) \bm{f},
    \label{equation:equations_of_motion}
\end{equation}
where $\bm{M}(\bm{q}) \in \mathbb{R}^{n_v \times n_v}$ is the mass matrix, and $\bm{h}(\bm{q}, \bm{v}) \in \mathbb{R}^{n_v}$ is the vector of Coriolis, centrifugal, and gravity terms.
On the right-hand side of the equation, $\bm{\tau} \in \mathbb{R}^{n_\tau}$ is the vector of joint torques commanded to the system,
and the selection matrix $\bm{S} = [\bm{0}_{n_\tau \times (n_v - n_\tau)} \quad \mathbb{I}_{n_\tau \times n_\tau}]$ selects which \gls{DoF} are actuated.
We consider that all limb joints are actuated, thus $n_\tau = n_j$.
The vector $\bm{\lambda} \in \mathbb{R}^{n_s}$ denotes the forces and torques\footnotemark\ experienced at the contact points, with $n_s$ being the total dimensionality of all contact wrenches.
The support Jacobian $\bm{J}_s \in \mathbb{R}^{n_s \times n_v}$ maps the contact wrenches $\bm{\lambda}$ to joint-space torques, and it is obtained by stacking the Jacobians which relate generalized velocities to limb end-effector motion as $\bm{J}_s = [\bm{J}_{C_1}^\top \quad \cdots \quad \bm{J}_{C_{n_c}}^\top]^\top$, with $n_c$ being the number of limbs in contact.
Finally, $\bm{f}$ represents any external force applied to the end-effector.
This force may be the result of a push or of some unpredicted disturbance.
Under nominal circumstances, this force is zero, i.e., $\bm{f} = \bm{0}$.
The Jacobian $J_e \in \mathbb{R}^{3 \times n_v}$ is used to map a linear force $\bm{f}$ applied at the end-effector to joint-space torques. %

\footnotetext{
    In the equations of motion above, $\bm{\lambda}$ represents the contact wrenches, i.e., forces and torques, acting on the robot.
    However, and similarly to \cite{dicarlo2018dynamic,winkler2018gait}, we chose to represent only the force-component as decision variables in our formulation.
    This is because the quadruped used in our experiments has point-like feet and we can assume the torque-component is negligible.
}

\subsection{Problem Discretization}
In order to plan motions for complex robot systems, we use an approach called \textit{direct transcription}, which is a powerful technique for \gls{TO}.

We start by converting the original motion planning problem (which is \emph{continuous} in time) into a numerical optimization problem that is \emph{discrete} in time.
We divide the trajectory into $N$ equally spaced segments,
$t_I = t_1 < \dots < t_M = t_F$,
where $t_I$ and $t_F$ are the start and final instants, respectively.
This division results in $M = N + 1$ discrete \textit{mesh points}, for each of which we explicitly discretize the states of the system, as well as the control inputs.
Let $x_k \equiv x(t_k)$ and $u_k \equiv u(t_k)$ be the values of the state and control variables at the $k$-th mesh point.
We treat $x_k \triangleq \{ \bm{q}_k, \bm{v}_k \}$ and $u_k \triangleq \{ \bm{\tau}_k, \bm{\lambda}_k \}$ as a set of \gls{NLP} variables, and formulate the basis of our trajectory optimization problem as
\begin{equation}
    \mathrm{find} \enspace \bm{\xi} \quad \mathrm{s.t.} \enspace x_{k+1} = f(x_k, u_k), \enspace x_k \in \mathcal{X}, \enspace u_k \in \mathcal{U},
    \label{equation:nlp}
\end{equation}
where $\bm{\xi}$ is the vector of decision variables,
$x_{k+1} = f(x_k, u_k)$ is the state transition function incorporating the nonlinear dynamics of the system,
and $\mathcal{X}$ and $\mathcal{U}$ are sets of feasible states and control inputs enforced by a set of equality and inequality constraints.
The decision variables vector $\bm{\xi}$ results from aggregating the generalized coordinates $\bm{q}_{1:M}$, generalized velocities $\bm{v}_{1:M}$, joint torques $\bm{\tau}_{1:N}$, and contact forces $\bm{\lambda}_{1:N}$, i.e.,
\begin{align}
    \bm{\xi} \triangleq \{ \bm{q}_1, \bm{v}_1, \bm{\tau}_1, \bm{\lambda}_1, \cdots, \bm{q}_{N}, \bm{v}_{N}, \bm{\tau}_{N}, \bm{\lambda}_{N}, \bm{q}_M, \bm{v}_M \}.
\end{align}

\subsection{System Constraints}
\label{subsec:system_constraints}
After having discretized the states and controls of the system over time as decision variables, we need to define a set of rules that restrict the motion represented by those variables.
We do this by specifying a set of mathematical equalities and inequalities, so that the solver ``knows'' how to compute trajectories that are not only physically feasible but that also complete the tasks we want the robot to solve.

\subsubsection{Domain of decision variables}
The most straightforward constraints we need to write are the lower and upper bounds of each decision variable in $\bm{\xi}$.
We constrain the joint positions, velocities, and torques to be within their corresponding lower and upper bounds.
\begin{align}
     & \bm{q}_L    \le \bm{q}_k    \le \bm{q}_U    &  & \forall k = 1 : M     \label{equation:bounds_position} \\
     & \bm{v}_L    \le \bm{v}_k    \le \bm{v}_U    &  & \forall k = 1 : M     \label{equation:bounds_velocity} \\
     & \bm{\tau}_L \le \bm{\tau}_k \le \bm{\tau}_U &  & \forall k = 1 : M - 1 \label{equation:bounds_torque}
\end{align}

\subsubsection{Initial and final velocities}
We enforce the initial and final velocities of every joint to be zero, i.e., $\bm{v}_1 = \bm{v}_M = \bm{0}$.
Note, however, that this is not a strict requirement of our framework but is chosen to ensure static start and end configurations.

\subsubsection{End-effector poses}
We enforce end-effector poses with
\begin{equation}
    f^\mathrm{fk}(\bm{q}_k, i) = \bm{p}_i,
    \label{equation:ee_poses}
\end{equation}
where $f^\mathrm{fk}(\cdot)$ is the forward kinematics function,
$\bm{q}_k$ are the joint coordinates at the $k$-th mesh point,
$i$ refers to the $i$-th end-effector of the robot, and
$\bm{p}_i \in SE(3)$ is the desired pose.
We use these constraints for defining the position and orientation of the robot's hand at specific mesh points, as well as to define the point contacts for the robot's feet during stance phases.\footnote{We do not constrain the robot's feet positions during leg swing phases.}
We pre-specify the contact sequence for the feet, which can be computed e.g. using contact planners such as \cite{tonneau2019sl1m}.

\subsubsection{Contact forces}
For mesh points where the robot is \emph{not} in contact with the environment (according to the pre-specified contact sequences), we enforce the contact forces at the respective contact points to be zero, i.e., $\bm{\lambda}_k = \bm{0}$.

\subsubsection{Friction cone constraints}
Similarly to \cite{caron2015leveraging}, we model friction at the contact points using an \textit{inner linear approximation} with a four-sided friction pyramid.
Consider the set of points $\{C_i\}$ where the robot is in contact with its environment.
Let $\bm{n}_i$ and $\mu_i$ be the unit normal and the friction coefficient of the support region at each contact, respectively.
A point contact remains fixed as long as its contact force $\bm{f}^c_i$ lies inside the linearized friction cone directed by $\bm{n}_i$:
\begin{align}
    | \bm{f}^{c}_i \cdot \bm{t}_i | & \leq (\mu_i / \sqrt{2}) (\bm{f}^{c}_i \cdot \bm{n}_i), \label{equation:friction_cone_1} \\
    | \bm{f}^{c}_i \cdot \bm{b}_i | & \leq (\mu_i / \sqrt{2}) (\bm{f}^{c}_i \cdot \bm{n}_i), \label{equation:friction_cone_2} \\
    \bm{f}^{c}_i \cdot \bm{n}_i     & > 0, \label{equation:friction_cone_3}
\end{align}
where $(\bm{t}_i, \bm{b}_i)$ form the basis of the tangential contact plane such that $(\bm{t}_i, \bm{b}_i, \bm{n}_i)$ is a direct frame.

\begin{figure*}[t]
    \captionsetup{font=small}
    \centering
    \includegraphics[width=\linewidth]{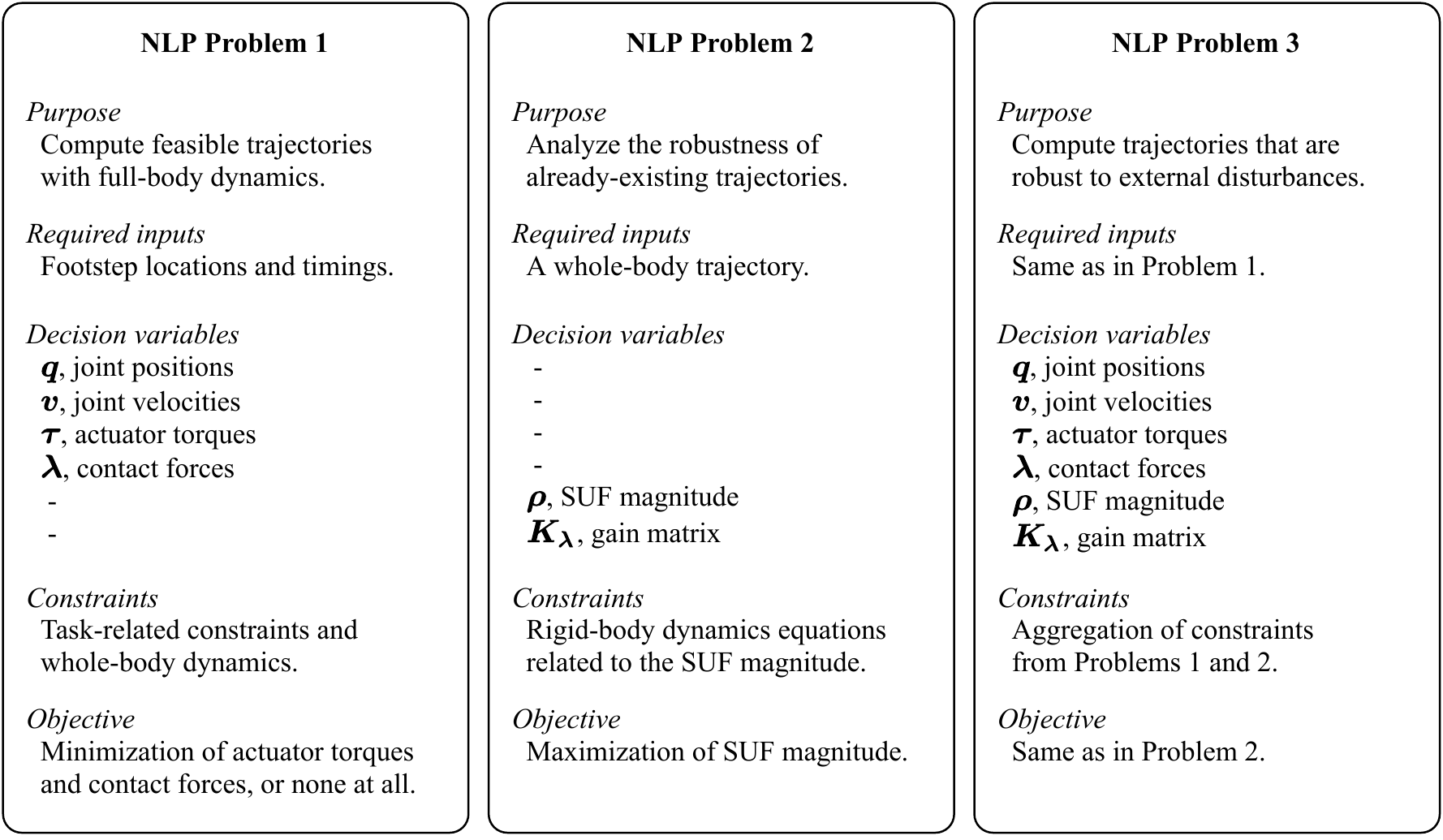}
    \caption{
        Block summaries of NLP Problems 1, 2, and 3.
        Inside each block, the summary states the purpose for using that formulation, the required inputs, the decision variables and constraints involved, and the objective function employed.
        The empty lines with a `-' under `Decision variables' emphasize that NLP Problem 3 is a combination of Problems 1 and 2.
    }\label{figure:problems}
\end{figure*}

\subsubsection{System dynamics}
We enforce the equations of motion (\autoref{equation:equations_of_motion}) using inverse dynamics.
This decision comes from our recent research on the benefits of inverse dynamics vs. forward dynamics for eliminating dynamics defects in direct transcription formulations.
In brief, we have shown that problems formulated using inverse dynamics are faster, more robust to coarser problem discretization, and converge in fewer iterations---see \cite{ferrolho2021inverse} for more details.

The inverse dynamics problem computes the joint torques and forces required to meet desired joint accelerations at a given state, i.e.,
\begin{equation}
    \bm{\tau}_{k}^{\ast} = f^\mathrm{id}(\bm{q}_{k}, \bm{v}_{k}, \bm{\dot{v}}_{k}^{\ast}, \bm{\lambda}_{k}),
    \label{equation:inverse_dynamics}
\end{equation}
where $f^\mathrm{id}(\cdot)$ is the function that solves the inverse dynamics problem,
and the desired joint accelerations can be calculated implicitly with
$\bm{\dot{v}}_{k}^{\ast} = (\bm{v}_{k+1} - \bm{v}_{k}) / h$.
We compute $\bm{\dot{q}}_{k+1}^{\ast}$ from $\bm{v}_{k+1}$, and integrate it (using semi-implicit Euler integration) to compute the next generalized coordinates $\bm{q}_{k+1}^{\ast}$.
Finally, we define the dynamics defect constraints as
\begin{equation}
    \bm{q}_{k+1}^{\ast} - \bm{q}_{k+1}   = \bm{0}
    \quad \text{and} \quad
    \bm{\tau}_{k}^{\ast} - \bm{\tau}_{k} = \bm{0}.
    \label{equation:defects_2}
\end{equation}

\subsection{Robustness Against Disturbances}

\subsubsection{NLP Problem 1}
Thus far, we have modeled the robot and its full body dynamics, discretized the motion planning problem, and defined a set of rules in the form of mathematical constraints.
At this stage, we have all the ``ingredients'' required for planning feasible trajectories that can be executed on the robot.
Henceforth, we will refer to this version of the formulation as \uline{NLP Problem 1}---a summary for this version of the formulation is shown on the left block in \autoref{figure:problems}.

Next, we are going to build upon our previous work~\cite{ferrolho2020optimizing} to present two different problem formulations, \uline{NLP Problem 2} and \uline{NLP Problem 3}, which can be used to analyze the robustness of known trajectories and to maximize the robustness of trajectories being computed, respectively.

\subsubsection{NLP Problem 2}
\label{subsubsec:nlp_problem_2}
When we compute a robot trajectory using NLP Problem 1, we may be interested in understanding how robust those trajectories are against forces applied to e.g. the end-effector.
Thus, one way of determining the robustness of that trajectory is by studying the set of forces that the end-effector is able to resist, both in terms of force magnitude and direction.
The metric proposed in our previous work \cite{ferrolho2020optimizing}, the \glsreset{SUF}\gls{SUF}, represents the smallest force magnitude (applied from any possible direction) that the robot is not able to counteract.
In other words, it gives the magnitude of the largest force that the robot can counteract in a worst-case scenario.
Next, we explain how to compute it.

Given a discretized robot trajectory (e.g., the output of NLP Problem 1), we can compute the \gls{SUF} magnitude throughout that motion by re-formulating the \gls{NLP} problem.
The decision variables for such a problem are
\begin{equation}
    \bm{\xi}_{\mathrm{NLP\ Problem\ 2}} \triangleq \{ \rho_1, {\overline{\bm{K}}_{\bm{\lambda}}}_1, \cdots, \rho_N, {\overline{\bm{K}}_{\bm{\lambda}}}_N \},
\end{equation}
where, for each $k$-th mesh point, $\rho_k$ is the magnitude of the \gls{SUF} and ${\overline{\bm{K}}_{\bm{\lambda}}}_k$ is the instantaneous gain matrix mapping a force expressed in end-effector space to ground-feet contact space.
Each and every $\rho_k$ is bound to $\mathbb{R^+}$, i.e.,
\begin{equation}
    \rho_k \ge 0 \quad \forall k = 1 : N.
\end{equation}
${\overline{\bm{K}}_{\bm{\lambda}}}_k$ have no explicit bounds; but they are constrained by
\begin{equation}
    \bm{a}_{\bm{\lambda}}^\top \bm{\lambda} + \left \| \bm{a}_{\bm{\lambda}}^\top \overline{\bm{K}}_{\bm{\lambda}} \right \| \le \bm{b}_{\bm{\lambda}},
    \label{equation:reformulation_constraint_3}
\end{equation}
with $\bm{a}_{\bm{\lambda}}$ and $\bm{b}_{\bm{\lambda}}$ pertaining to the alternative form of writing the friction cone constraints, i.e., $\bm{A}_{\bm{\lambda}} \bm{\lambda} \le \bm{b}_{\bm{\lambda}}$.\footnote{We refer readers to \cite{ferrolho2020optimizing} for a full explanation and derivation of the terms.}
Akin to the equations of motion, the relationship between the \gls{SUF} and the robot capabilities at every mesh point is given by
$\bm{S}^\top \overline{\bm{K}}_{\bm{\tau}} + \bm{J}_s^\top \overline{\bm{K}}_{\bm{\lambda}} + \bm{J}_e^\top \rho = \bm{0}$.
And this can be rewritten in a way that highlights the inherent structure of the constraint:
\begin{equation}
    \begin{bmatrix} \bm{0} \\ \mathbb{I} \end{bmatrix} \overline{\bm{K}}_{\bm{\tau}} = - \begin{bmatrix} \bm{J}_s^{\top_\mathrm{base}} \\ \bm{J}_s^{\top_\mathrm{limbs}} \end{bmatrix} \overline{\bm{K}}_{\bm{\lambda}} - \begin{bmatrix} \bm{J}_e^{\top_\mathrm{base}} \\ \bm{J}_e^{\top_\mathrm{limbs}} \end{bmatrix} \rho,
    \label{equation:constraint_splittage}
\end{equation}
where ${\overline{\bm{K}}_{\bm{\tau}}}$ is an instantaneous gain matrix mapping a force expressed in the end-effector frame to joint-torque space.
We enforce this relationship by splitting it into two parts.
For the top part of the equation, concerning the floating base, we write the following nonlinear equality:
\begin{equation}
    \bm{J}_s^{\top_\mathrm{base}} \overline{\bm{K}}_{\bm{\lambda}} + \bm{J}_e^{\top_\mathrm{base}} \rho = \bm{0}.
    \label{equation:transcribed_constraint_1}
\end{equation}
As for the bottom part, concerning the limbs of the robot, we write the following nonlinear inequality:
\begin{equation}
    \bm{a}_{\bm{\tau}}^\top \bm{\tau} + \left \| \bm{a}_{\bm{\tau}}^\top \left ( - \bm{J}_s^{\top_\mathrm{limbs}} \overline{\bm{K}}_{\bm{\lambda}} - \bm{J}_e^{\top_\mathrm{limbs}} \rho \right ) \right \| \le \bm{b}_{\bm{\tau}}.
    \label{equation:transcribed_constraint_2}
\end{equation}

Once we have defined the above constraints and decision variables, we can maximize the following objective function with any off-the-shelf nonlinear solver:
\begin{equation}
    \argmax_{\bm{\xi}_{\mathrm{NLP\ Problem\ 2}}} \quad \sum_{k=1}^{N} \, \rho_k.
    \label{equation:objective_max_rho}
\end{equation}

In summary, for a given (constant) trajectory, the outcome of this nonlinear optimization problem will be the magnitude of the \gls{SUF} over time (i.e., the $\rho_k$ for every mesh point of the discretized trajectory) and ${\overline{\bm{K}}_{\bm{\lambda}}}_k$.
$\overline{\bm{K}}_{\bm{\tau}}$ are also an output, since they can be computed as a function of $\overline{\bm{K}}_{\bm{\lambda}}$ and $\rho$ without performing any inversion---as hinted by \autoref{equation:constraint_splittage}.
The outputs ${\overline{\bm{K}}_{\bm{\lambda}}}$ and $\overline{\bm{K}}_{\bm{\tau}}$ can be used to understand and explain the specific constraint that determines the upper bound of the \gls{SUF} (i.e., friction cone or torque, on which foot or motor), although that is something we do not investigate in this paper.

\subsubsection{NLP Problem 3}
NLP Problems 1 and 2 allow us to compute a feasible whole-body trajectory and to calculate the robustness of said trajectories, respectively.
By combining NLP Problem 1 and NLP Problem 2, we obtain a single (albeit more complex) problem formulation which is able to compute whole-body trajectories that are not only feasible but also more robust against external disturbances.
We call this single formulation \emph{NLP Problem 3}, and it is summarized on the right block in \autoref{figure:problems}.
The decision variables of NLP Problem 3 are
\begin{equation}
    \begin{aligned}
        \bm{\xi}_{\mathrm{NLP\ Problem\ 3}} \triangleq \{ & \bm{q}_1, \bm{v}_1, \bm{\tau}_1, \bm{\lambda}_1, \rho_1, {\overline{\bm{K}}_{\bm{\lambda}}}_1,         \\
                                                          & \cdots,                                                                                                \\
                                                          & \bm{q}_{N}, \bm{v}_{N}, \bm{\tau}_{N}, \bm{\lambda}_{N}, \rho_N, {\overline{\bm{K}}_{\bm{\lambda}}}_N, \\
                                                          & \bm{q}_M, \bm{v}_M \}.
    \end{aligned}
\end{equation}
The constraints are the combined constraints of NLP Problems 1 and 2.
The objective function is the same as NLP Problem 2, i.e., the maximization of every mesh point's \gls{SUF} added up:
\begin{equation}
    \argmax_{\bm{\xi}_{\mathrm{NLP\ Problem\ 3}}} \quad \sum_{k=1}^{N} \, \rho_k.
\end{equation}

\subsection{Contact Switching}
In contrast to our previous work \cite{ferrolho2020optimizing}, the \gls{NLP} formulations in \autoref{figure:problems} consider the making and breaking of contacts between the feet of the robot and its environment.
This is one of the main contributions of this paper, and we will now explain how we have enabled this.

As we have previously explained in `System Constraints' (\autoref{subsec:system_constraints}), during problem discretization, we handle feet that are in contact with the ground (\emph{stance phase}) differently from feet that are moving through free space (\emph{swing phase}).
In short, for each mesh point of the trajectory, and for each foot of the robot:
if that foot is in stance phase, we enforce it to remain still and we also enforce the contact force to lie within friction cone boundaries; otherwise, we enforce only a zero contact force constraint.
Since this is done during the problem transcription process, it requires knowing the timings for contact switches \textit{a priori}, as well as feet positions\footnote{For now, we will consider the feet positions to be fixed, but we will release this limitation in \autoref{sec:contact_location_optimization}, where we discuss this subject further.} for each stance phase.
The upside is that the problem complexity does not increase much, especially compared with other approaches that consider contact switching by formulating complementarity constraints (e.g. Posa \textit{et al.} \cite{posa2014direct}).

The approach explained in the last paragraph is enough for enabling contact switching in NLP Problem 1.
However, additional changes are needed for Problems 2 and 3, since those problem formulations involve extra constraints and decision variables regarding the maximization of the \gls{SUF}.
In essence, the decision variables ${\overline{\bm{K}}_{\bm{\lambda}}}_k$ pertaining to the feet in swing phase must be set to zero, since no contact forces exist in that context.
Fortunately, we need not worry about $\overline{\bm{K}}_{\bm{\tau}_k}$, since it is defined as a function of ${\overline{\bm{K}}_{\bm{\lambda}}}_k$ (see \autoref{equation:constraint_splittage}).
Nonetheless, the most challenging aspect of these modifications is not in the writing of the constraints, but rather in the writing of the functions that evaluate the Jacobian of those constraints (together with their sparsity structure).
To this end, we use modern automatic differentiation capabilities from Julia \cite{bezanson2017julia}, but the process of passing the results of the Jacobian and sparsity structure evaluations to specialized \gls{NLP} solvers remains tricky and requires particular attention by the programmer to the indexing of the decision variables and constraints.
We formulate large but sparse \gls{NLP} problems through our framework and therefore specifying the sparsity pattern is very important to attain shorter computation times, as the Jacobian of the constraints contain mostly zeroes (especially the dynamics Jacobian, which is block diagonal).

\subsection{Computation Time}
It is not trivial to formulate or solve the optimization problems described in this paper.
Special care needs to be taken when transcribing the constraints and objective functions, otherwise computation becomes prohibitively expensive---especially when considering full body dynamics models.
Here, we show that our approach is computationally feasible as an offline planning tool suitable for industrial applications.

The following evaluations were carried out in a single-threaded process on a MacBook Air (2020) with an Apple M1 chip and \SI{8}{\giga\byte} of unified memory.
Our framework has been implemented in Julia~\cite{bezanson2017julia},
using the rigid-body dynamics library \gls{RBD.jl}~\cite{rigidbodydynamicsjl}, and the Artelys Knitro~\cite{byrd2006knitro} optimization library.
We used the interior-point method of Waltz \textit{et al.} \cite{waltz2006interior} to solve the \gls{NLP} problems.
Automatic sparsity detection was used to provide sparsity patterns to the solver, and automatic differentiation (forward mode) for evaluating the Jacobians of the constraints---but we allowed Knitro to compute the Hessians via the limited-memory quasi-Newton BFGS method.

\begin{table}[ht]
    \captionsetup{font=small}
    \centering
    \caption{Average computation time (in seconds) for solving the valve task at different discretization rates (\SI{50}{\hertz} and \SI{100}{\hertz}).}
    \label{table:computation_times}
    \begin{tabular}{
            c
            S[table-format=2.2(3), separate-uncertainty=true]
            S[table-format=2.1(2), separate-uncertainty=true]
        }
        \toprule
                      & \SI{50}{\hertz} & \SI{100}{\hertz} \\
        \midrule
        NLP Problem 1 & 1.69(42)        & 7.5(84)          \\
        NLP Problem 2 & 12.40(590)      & 23.7(73)         \\
        NLP Problem 3 & 26.00(1500)     & 38.0(240)        \\
        \bottomrule
    \end{tabular}
\end{table}

In \autoref{table:computation_times}, we present the elapsed time (in seconds) for solving the three NLP problems summarized in \autoref{figure:problems}, for an industrial task where the quadruped robot must rotate a wheel/valve using the arm mounted on its torso.\footnote{
    Apart from this example, an in-depth evaluation of the NLP performance is beyond the scope of this paper. For further investigation, refer to \cite{ferrolho2020optimizing}.
}
The average solving time was determined by solving each problem multiple times with different initial conditions. These conditions involved random translations and rotations of the robot's base relative to the manipulated object.

Based on the table, it is evident that tackling these problems involves a high computational cost, making our approach impractical for online planning.
However, the computation times are still quite reasonable to consider using our framework offline, enabling us to create a repository of behaviors that can be efficiently accessed during online execution.

\begin{table}[ht]
    \captionsetup{font=small}
    \centering
    \caption{Characteristics of the NLP Problems for a \SI{4}{\second}-long trajectory discretized at \SI{100}{\hertz} (valve task).}
    \label{table:nlp_characteristics}
    \begin{tabular}{
            lrrr
        }
        \toprule
                               & NLP Prob 1 & NLP Prob 2 & NLP Prob 3 \\
        \midrule
        Decision Variables     & $ 32\,448$ & $ 14\,800$ & $ 47\,248$ \\
        Linear Inequalities    & $  8\,000$ & $       0$ & $  8\,000$ \\
        Nonlinear Equalities   & $ 25\,218$ & $  7\,200$ & $ 32\,418$ \\
        Nonlinear Inequalities & $       0$ & $ 22\,400$ & $ 22\,400$ \\
        Nonzeros in Jacobian   & $548\,538$ & $237\,600$ & $993\,738$ \\
        \bottomrule
    \end{tabular}
\end{table}

In relation to the size of the problems, employing a \SI{100}{\hertz} discretization over a trajectory duration of \SI{4}{\second} yields a total of 401 knot points.
\autoref{table:nlp_characteristics} shows the number of decision variables, constraints, and nonzeros in the Jacobian for that amount of knot points, offering some insight into the complexity of problems of such size.

\section{Experiments and Results}
We now present the experiments we carried out for evaluating our work and their respective results.
This section is organized as follows:
\begin{itemize}
    \item [\textit{A.}] Describes the system integration work required for combining the quadruped robot and the robot arm, as well as our planning framework with existing controllers;
    \item [\textit{B.}] Presents a repeatability test where we commanded the robot to turn an industrial hand wheel multiple times from different starting positions and orientations;
    \item [\textit{C.}] Compares the robustness of two distinct trajectories for turning a hand wheel and for pulling a lever;
    \item [\textit{D.}] Explains how the \gls{SUF} can be used as a tool for analyzing existing robot trajectories;
    \item [\textit{E.}] Demonstrates the capabilities of our framework to plan robust whole-body motions involving making and breaking of contacts; and finally,
    \item [\textit{F.}] Tests the robustness of a loco-manipulation trajectory for lifting a bucket with incremental weights until failure.
\end{itemize}

Subsections \textit{A}, \textit{B}, \textit{C}, and \textit{F} are concrete evaluations of our method on real robot hardware.
Whereas the focus of subsections \textit{D} and \textit{E} is to demonstrate new features.

\subsection{System Integration}
In order to demonstrate the capabilities of our planning framework on the real robot, we integrated it with the existing software stacks of the ANYmal quadruped and Kinova arm.

For the quadruped, we used one of ANYbotics' software releases which comes with multiple locomotion controllers working out-of-the-box. Human operators can control the robot remotely via a joystick to tell the robot where to walk and which gait to use, or they can pre-specify a mission as a set of waypoints for the robot to walk through.
Moreover, they provide an interface for specifying a custom payload attached to the robot, but this is assumed to be a static payload, such as a thermal camera, or an imaging sensor.

For controlling the robot as a whole (quadruped\,+\,arm), we settled on two operation modes: \textsl{teleoperated} and \textsl{autonomous}.
When the robot is in \textsl{teleoperated mode}, we do not move the Kinova arm; it remains still in a ``parked'' configuration (stowed on top of the quadruped).
Doing this allows us to consider the arm a static payload, which we can specify using the interface provided by ANYbotics.
In turn, this allows us to take advantage of every existing capability provided by ANYbotics' stack.
Finally, whenever we wish to operate the arm, we switch to \textsl{autonomous mode}.
In this mode, we use a custom controller\footnotemark\ for the quadruped base and Kinova's velocity controller for the arm.
This is the mode we use to execute the whole-body trajectories generated with our planning framework.

\footnotetext{
    We use the same controller as in our previous work \cite{ferrolho2020optimizing}.
    We commanded each joint of the quadruped with feedforward torque and feedback on position and velocity.
    Position, velocity, and torque references are updated at \SI{400}{\hertz}.
}

In summary, we start the robot in \textsl{teleoperated mode} by default. In this mode, we can walk the robot on flat ground and over ramps, but we cannot move the arm. We use the \textsl{teleoperated mode} to walk the robot around a facility and towards points of interest.
Then, once the robot reaches said points of interest, it awaits an instruction (e.g, ``turn the wheel'', or ``pull the lever'').
As soon as the robot is given an instruction, it switches into \textsl{autonomous mode}, our planning framework is triggered and computes a whole-body trajectory for the robot. This trajectory is then passed on to the arm and quadruped controllers for synchronous execution.
Once the robot finishes executing the task autonomously, it switches back into \textsl{teleoperated mode} and the system returns control to the human operator, who can walk the robot remotely towards the next task of the mission.

A demonstration of the entire system is available here: \texttt{\small\url{https://youtu.be/3qXNHVCagL8}}.
In the video, a human teleoperates the robot to walk around a mock-up scaffolding of an offshore platform (such as an oil rig). The operator approaches different points of interest, and commands the robot to autonomously turn a hand wheel, pull a lever, push a gate whilst standing on a ramp, and pull a rope to lift a \SI{1.1}{\kilo\gram} bucket.
These tasks are the ones shown in \autoref{figure:cover}.
Although we have focused on tasks relevant for industrial inspection, our framework allows us to formulate virtually any loco-manipulation task through the definition of a set of constraints for the robot.

\begin{figure}[t]
    \captionsetup{font=small}
    \centering
    \begin{subfigure}[t]{\linewidth}
        \includegraphics[width=0.498\linewidth]{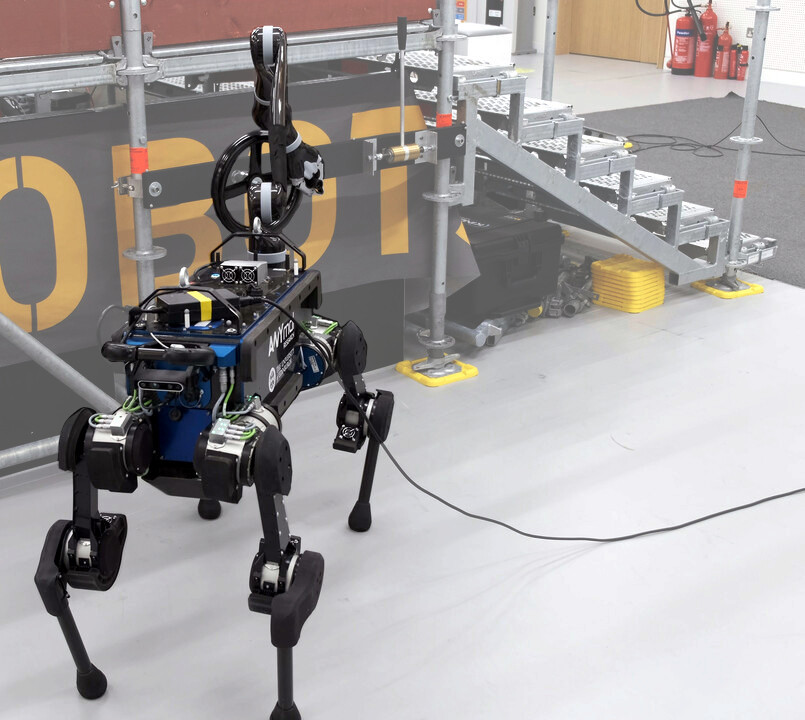}\hfill%
        \includegraphics[width=0.498\linewidth]{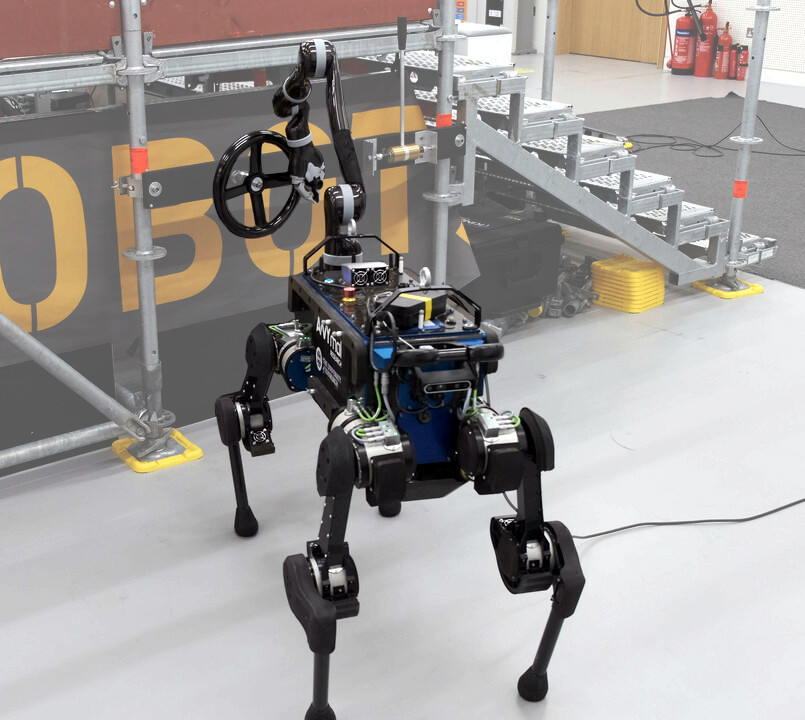}\hfill%
    \end{subfigure}\hfill%
    \vspace{0.004\linewidth}
    \begin{subfigure}[t]{\linewidth}
        \includegraphics[width=0.498\linewidth]{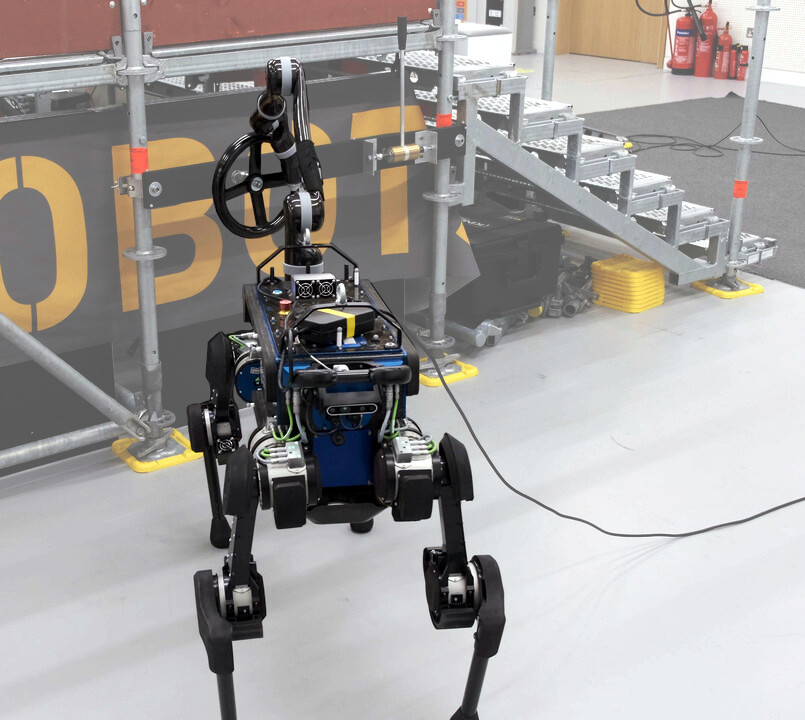}\hfill%
        \includegraphics[width=0.498\linewidth]{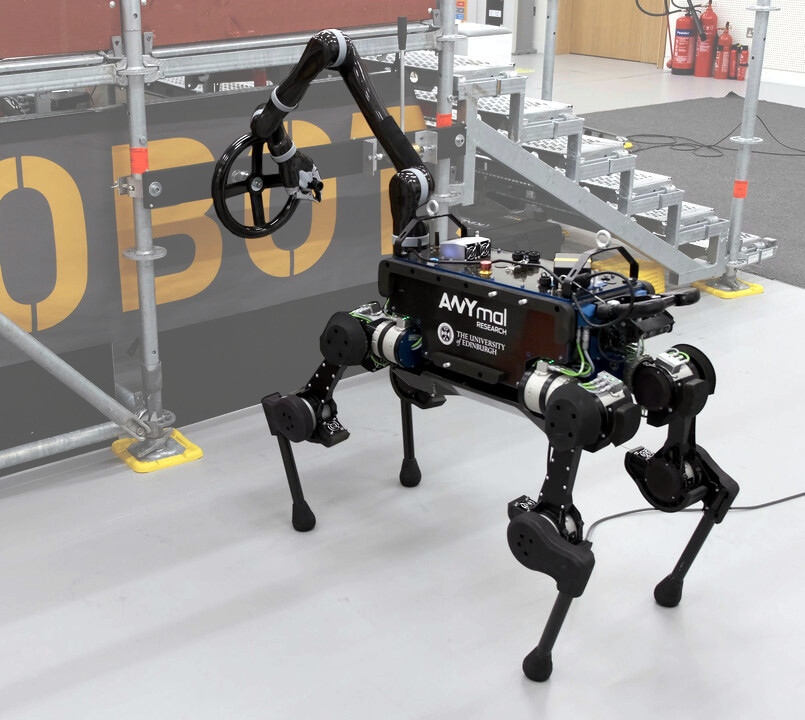}\hfill%
    \end{subfigure}\hfill%
    \vspace{0.004\linewidth}
    \begin{subfigure}[t]{\linewidth}
        \includegraphics[width=0.498\linewidth]{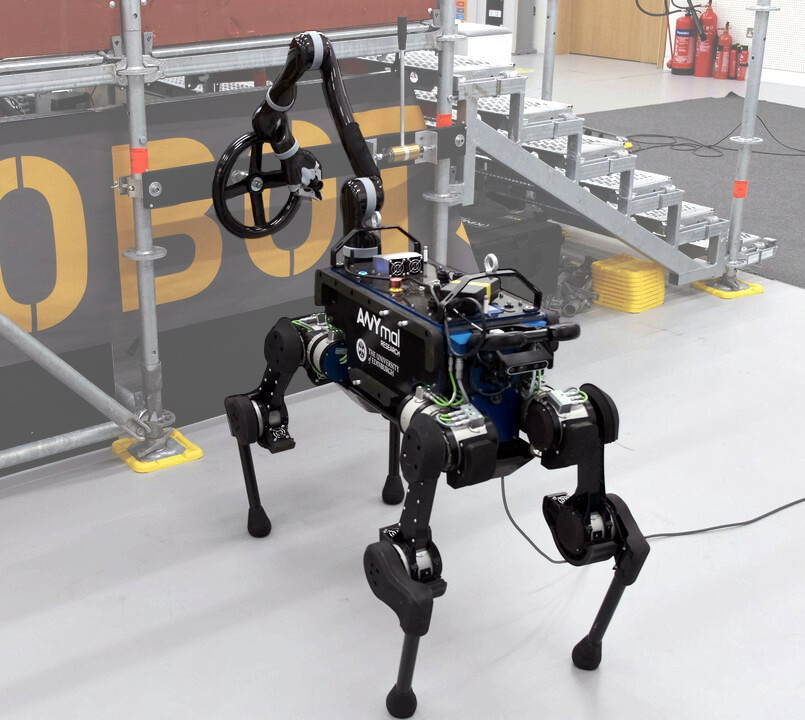}\hfill%
        \includegraphics[width=0.498\linewidth]{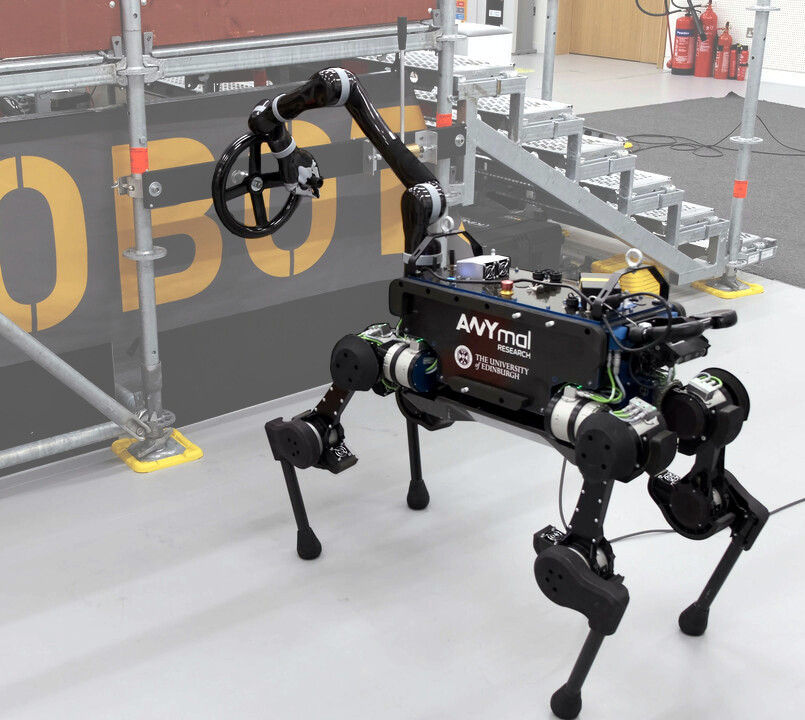}\hfill%
    \end{subfigure}
    \caption{
        Snapshots of the robot grasping the hand wheel during the repeatability test.
        Notice how the position and orientation of the robot base relative to the wheel are different on every snapshot.
        Video: \texttt{\small\url{https://youtu.be/Ok8Pcwn_I0w}}.
    }\label{figure:repeatability_test}
\end{figure}

\subsection{Repeatability Test}
Our goal for this experiment was to ensure the following characteristics of our system:
\begin{itemize}
    \item the robot is able to operate continuously for extended periods of time without falling;
    \item the operator is able to send walking commands to the robot during \textsl{teleoperated mode}, but not during \textsl{autonomous mode};
    \item the Vicon motion capture system calculates the relative transform between the free-floating base of the robot and the manipulation target reliably;
    \item the planning framework computes a feasible whole-body trajectory for completing the manipulation task;
    \item the controller is able to reach and grasp targets, and track reference trajectories accurately.
\end{itemize}

To verify the points above, we carried out a repeatability test for turning an industrial hand wheel.
During this test, we operated the robot continuously for \SI{30}{\minute} without a safety harness.
During this time, a human operator walked the robot to different points near the manipulation target and triggered the ``turn wheel'' behaviour.
The test was completed successfully and our system passed all of the points listed above.
\autoref{figure:repeatability_test} shows a few snapshots of the robot grasping the wheel during the test.
Video footage is also available here: \texttt{\small\url{https://youtu.be/Ok8Pcwn_I0w}}.

\subsection{SUF Optimization for Turning a Wheel and Pulling a Lever}
\begin{figure}[t]
    \captionsetup{font=small}
    \begin{subfigure}[t]{\linewidth}
        \includegraphics[width=0.498\linewidth]{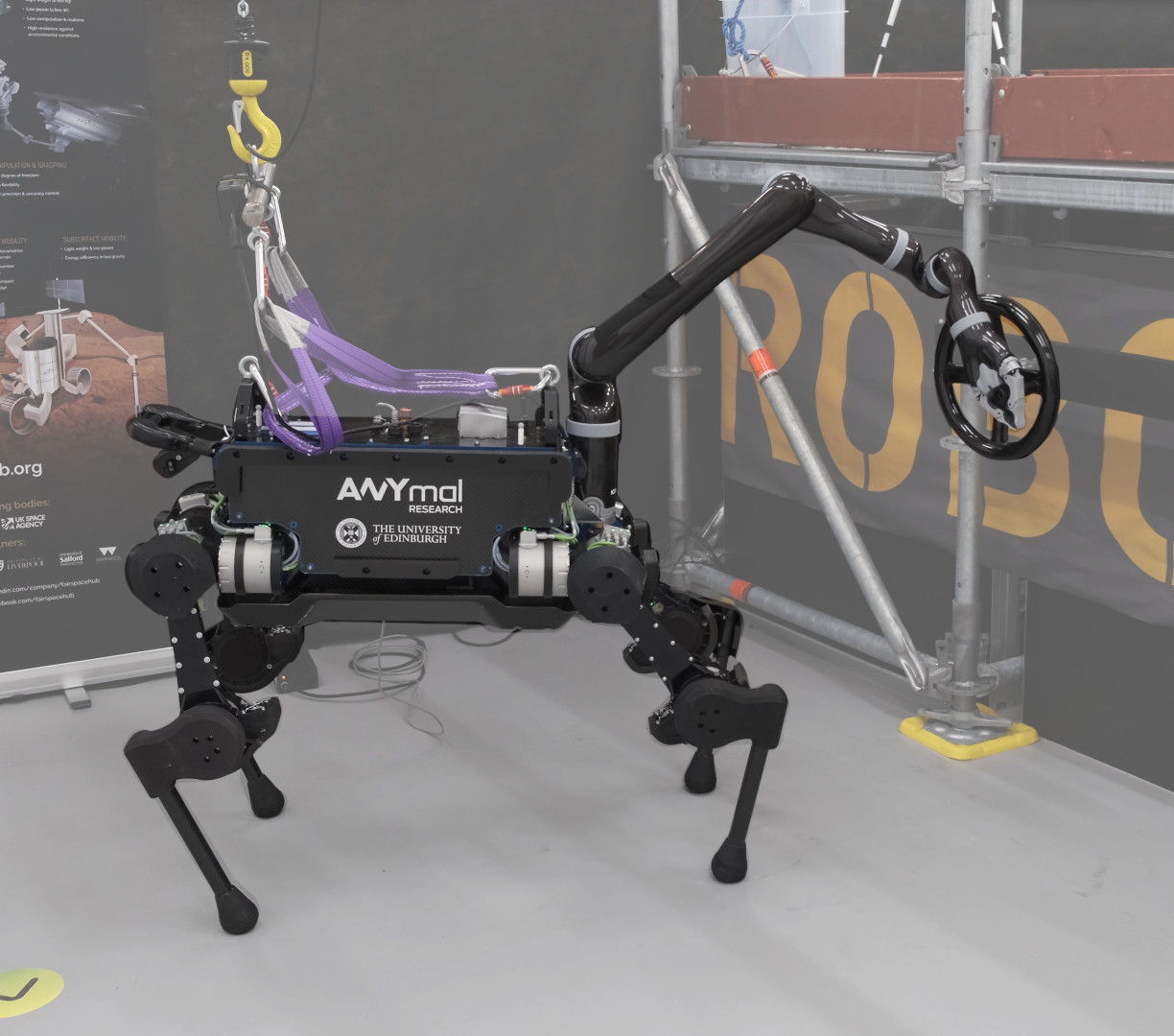}\hfill%
        \includegraphics[width=0.498\linewidth]{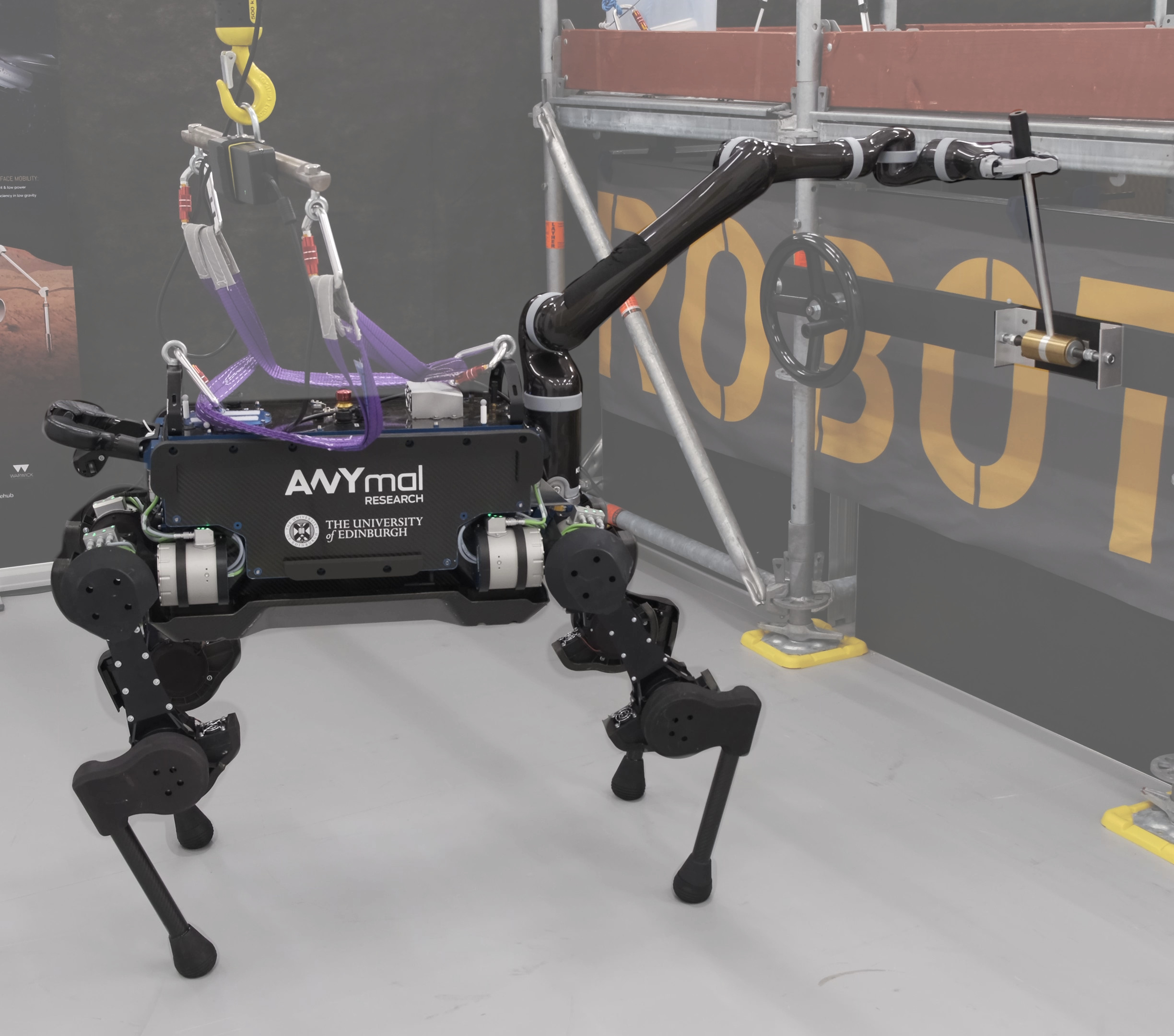}\hfill%
        \caption{Baseline}
    \end{subfigure}\hfill%
    \vspace{0.5em}
    \begin{subfigure}[t]{\linewidth}
        \includegraphics[width=0.498\linewidth]{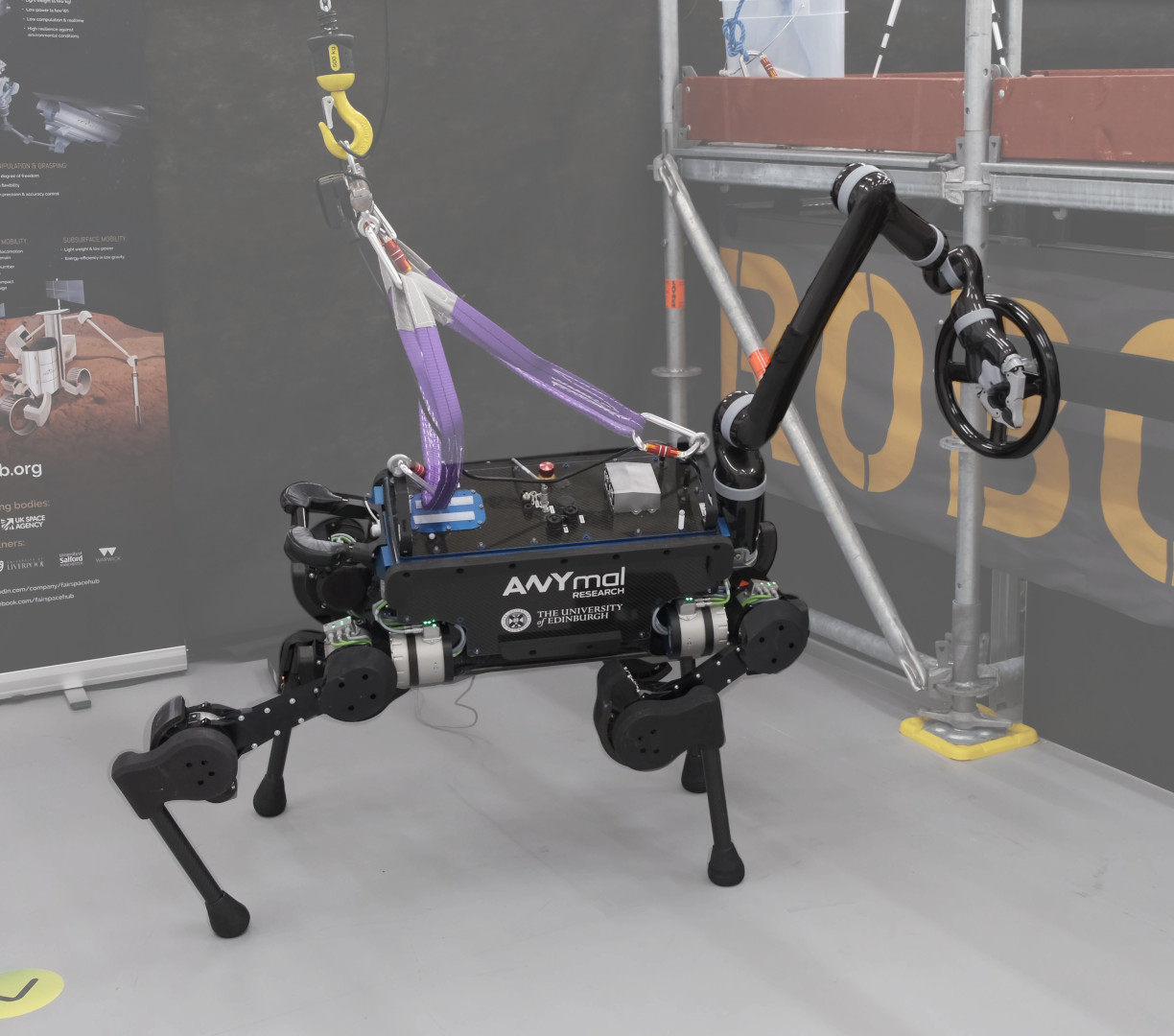}\hfill%
        \includegraphics[width=0.498\linewidth]{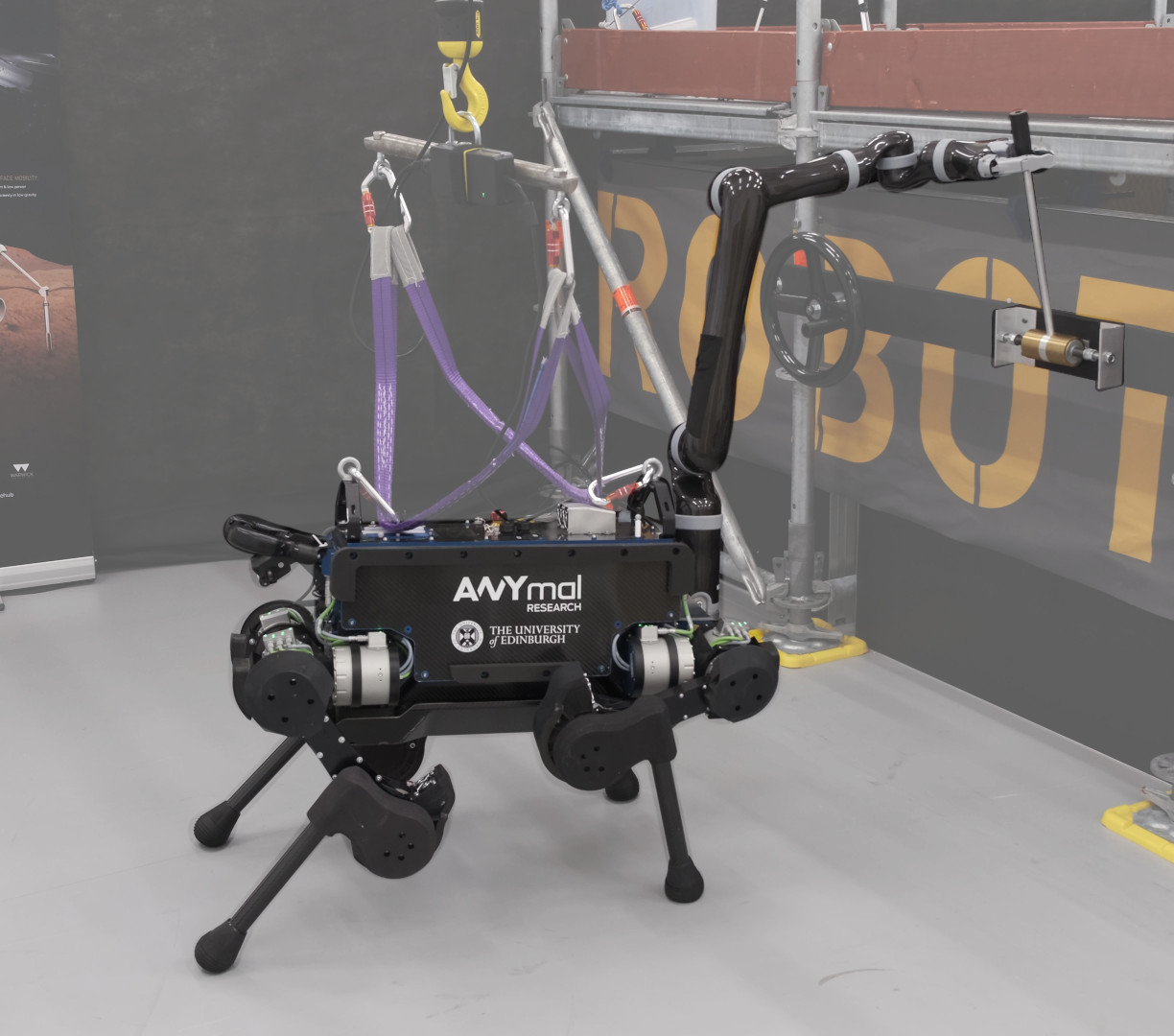}\hfill%
        \caption{Proposed}
    \end{subfigure}\hfill%
    \vspace{0.5em}
    \begin{subfigure}[t]{\linewidth}
        \includegraphics[width=0.498\linewidth]{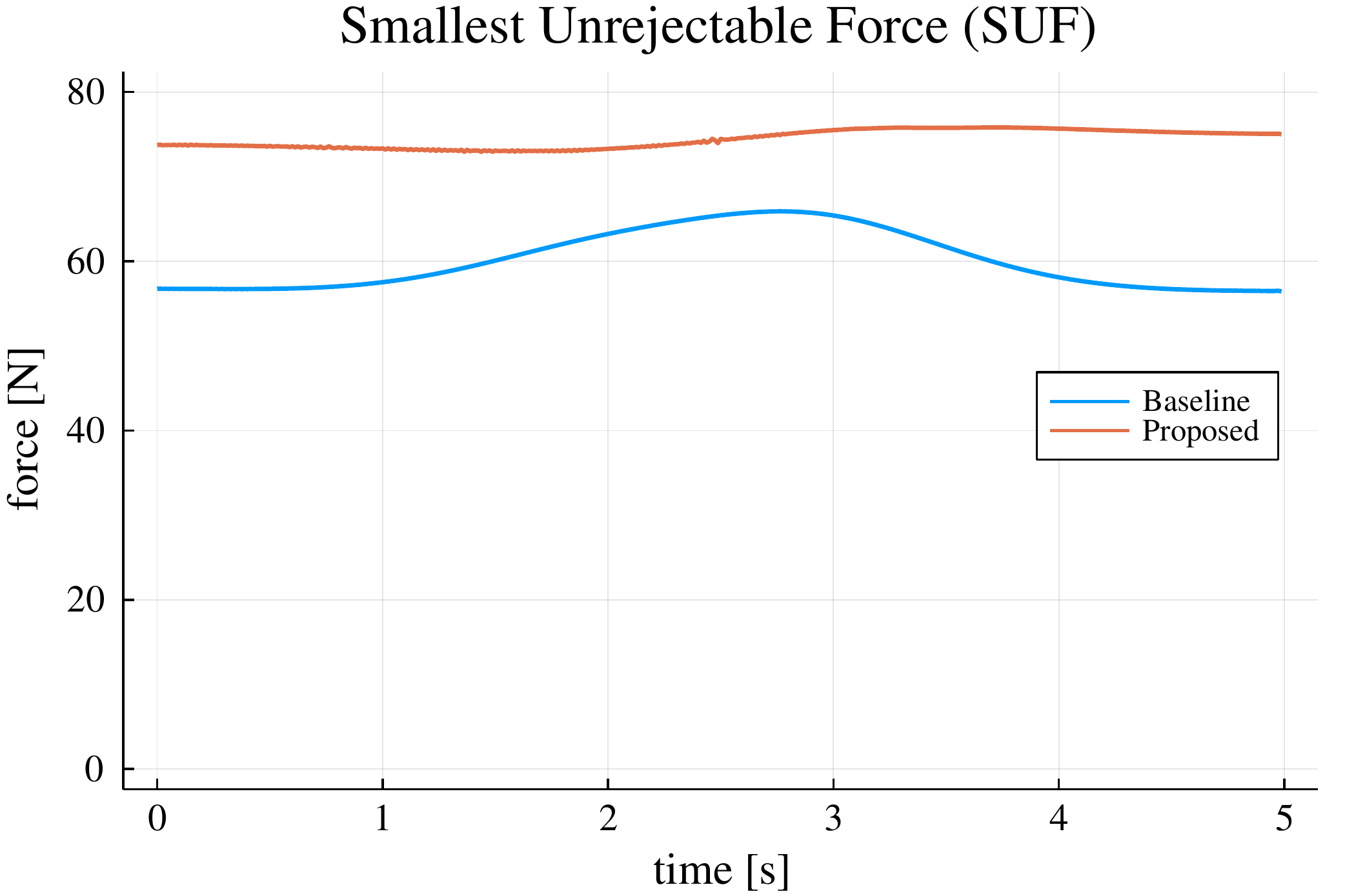}\hfill%
        \includegraphics[width=0.498\linewidth]{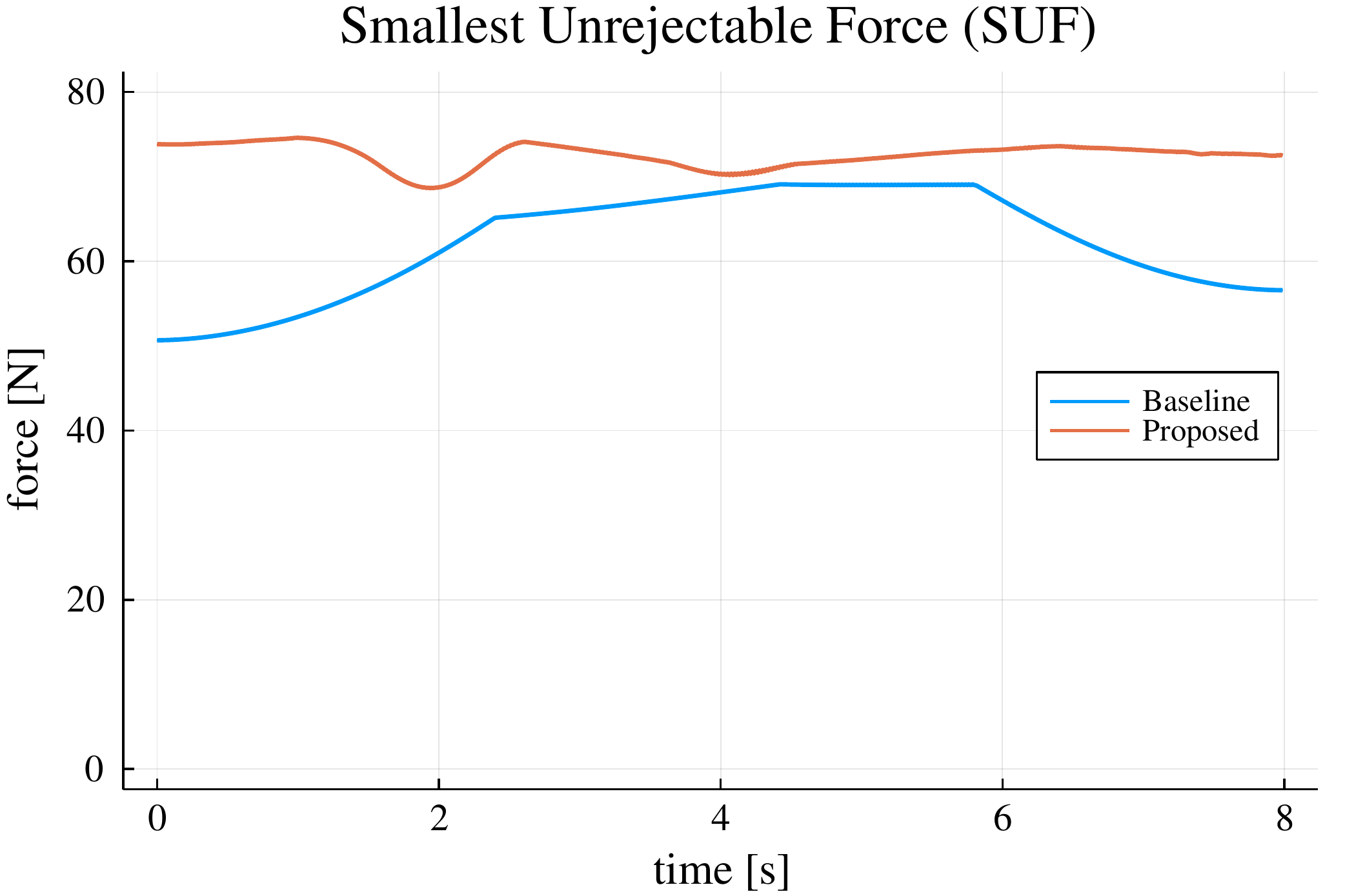}\hfill%
        \caption{\gls{SUF} magnitude over time}
    \end{subfigure}
    \caption{
        From top to bottom: initial configuration of the \textsl{baseline} trajectories (top), initial configuration of the \textsl{proposed} trajectories (middle), and plot comparing the \gls{SUF} of the \textsl{baseline} and \textsl{proposed} trajectories (bottom).
        Left and right columns concern the tasks of turning an industrial hand wheel and pulling down a lever, respectively.
    }\label{figure:suf_comparison}
\end{figure}

In this experiment, our goal is to compare the resulting \gls{SUF} of two different objective functions:
the first trajectory (\textsl{baseline}) is optimized considering a cost function that minimizes torques and contact forces,
while the second trajectory (\textsl{proposed}) maximizes the \gls{SUF} explicitly.
We run this experiment for two tasks:
(i) turn a hand wheel clockwise by a full revolution, and
(ii) pull down a lever from its resting position.
\autoref{figure:suf_comparison} shows an instance of the results of this experiment.

\subsubsection{Turning the hand wheel}
In \autoref{figure:suf_comparison}'s left column, we can see that the \gls{SUF} magnitude of the proposed trajectory (orange line) is within \qtyrange{73}{76}{\newton} throughout the entire motion;
whereas the \gls{SUF} of the baseline trajectory (blue line) lies within \qtyrange{56}{66}{\newton}---it starts and ends at ${\sim}\SI{56}{\newton}$, increasing slightly in the middle of the trajectory where it peaks at ${\sim}\SI{66}{\newton}$.
The \gls{SUF} mean-percentage-increase of the proposed approach versus the baseline is approximately \SI{24}{\percent}.
Video: \texttt{\small\url{https://youtu.be/1M32AHuuDhI}}.

\subsubsection{Pulling the lever}
In \autoref{figure:suf_comparison}'s right column, the \gls{SUF} magnitude of the proposed trajectory (orange line) remains within \qtyrange{68}{75}{\newton} throughout the entire motion;
on the other hand, the \gls{SUF} magnitude of the baseline trajectory (blue line) remains within \qtyrange{50}{69}{\newton}.
The \gls{SUF} mean-percentage-increase of the proposed approach versus the baseline is approximately \SI{18}{\percent}.
Video: \texttt{\small\url{https://youtu.be/6A9eSdfcj7A}}.

We repeated this experiment multiple times for different robot orientations relative to the wheel and the lever.
The results were identical to those shown in \autoref{figure:suf_comparison}.
Both the baseline and proposed trajectories successfully completed the task.
However, trajectories planned using the proposed approach outperformed the baseline version in both tasks in terms of their robustness against external disturbances.

\subsection{The SUF as a Tool for Analyzing Trajectories}
Earlier in this manuscript, when we presented \autoref{figure:problems}, we explained how the \gls{SUF} allows us to analyze the robustness of existing trajectories, and also how it allows us to compare multiple trajectories with each other.
In this experiment, we want to further develop the idea of using the \gls{SUF} as an analysis tool.
Our goal is to show that, thanks to our problem formulation, we can analyze individual trajectories to understand e.g. how important of a role each of the legs play during a motion.

\begin{figure}[t]
    \captionsetup{font=small}
    \centering
    \begin{subfigure}[t]{\linewidth}
        \includegraphics[width=0.7\linewidth,center]{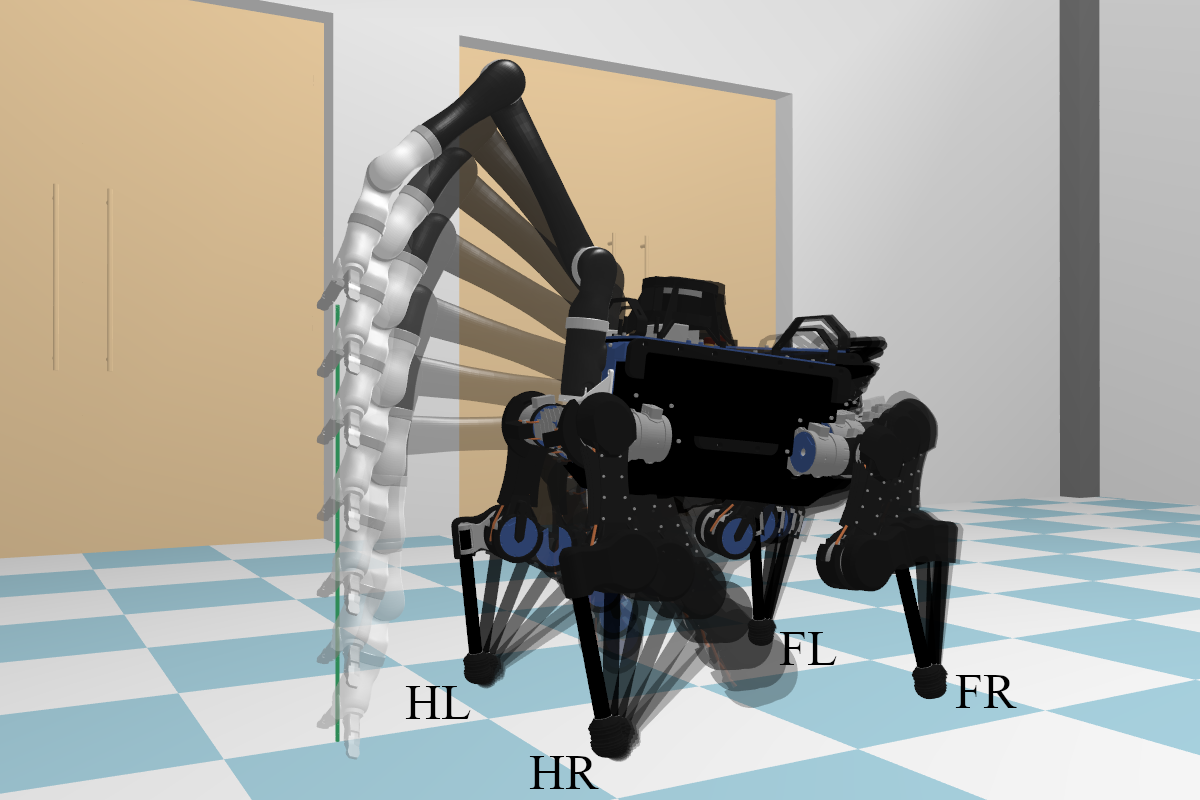}
        \caption{}\label{figure:suf_leg_analysis_top}
    \end{subfigure}\hfill%
    \begin{subfigure}[t]{\linewidth}
        \includegraphics[width=0.8\linewidth,center]{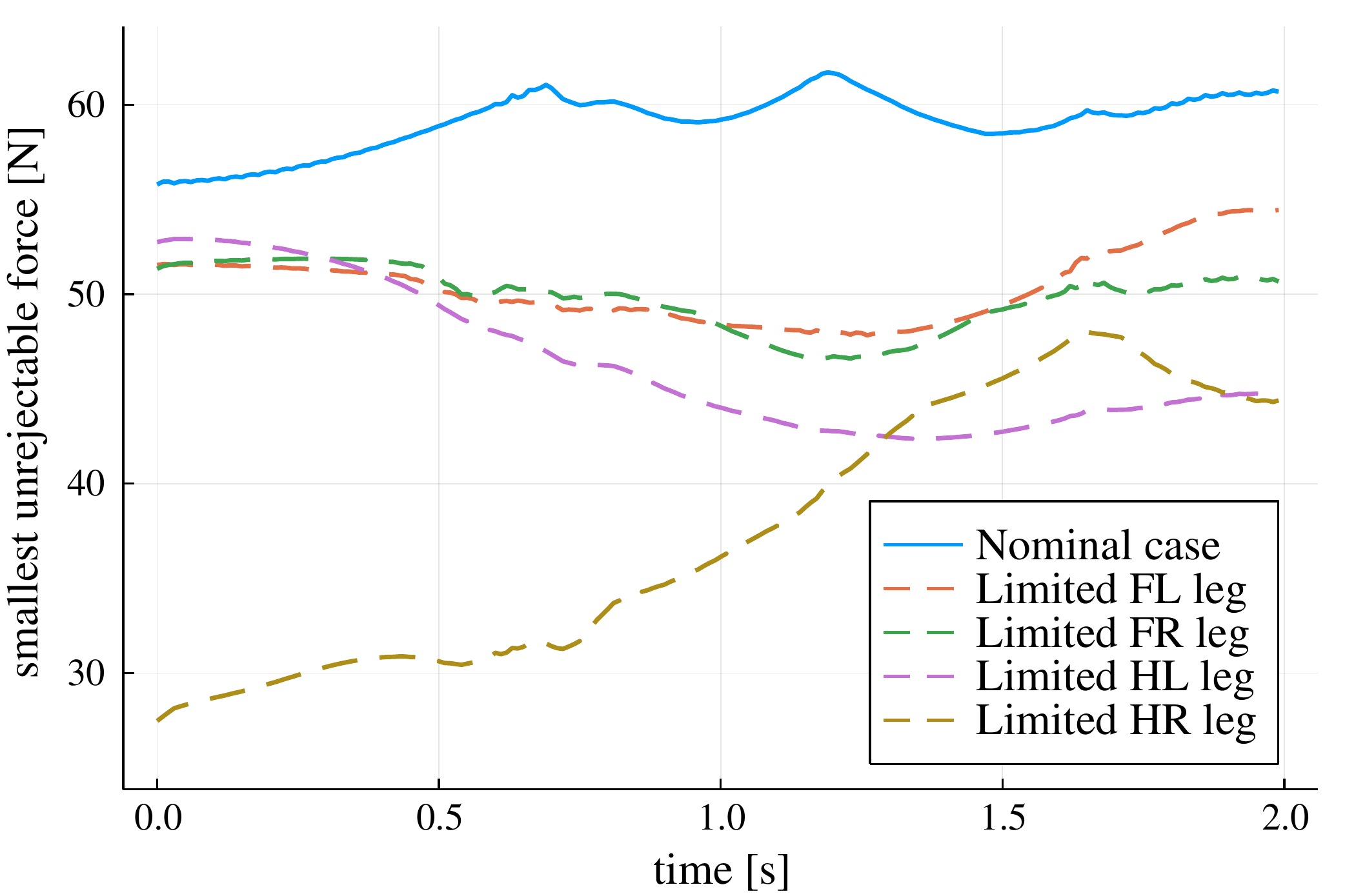}
        \caption{}\label{figure:suf_leg_analysis_bottom}
    \end{subfigure}
    \caption{
        The image on the top shows a motion plan for picking up an object from the floor.
        The plot on the bottom shows, for each leg, how the \gls{SUF} decreases if we were to consider the current torques to be at actuator limits and current contact force at the friction cone boundary.
    }\label{figure:suf_leg_analysis}
\end{figure}

In \autoref{figure:suf_leg_analysis_top}, we show a robot trajectory planned with our framework for picking up something from the ground;
and in \autoref{figure:suf_leg_analysis_bottom}, the solid blue line represents the \gls{SUF} over time for that motion.
An important question we may ask is: how important is each leg for a successful execution of the motion?
We can ask a similar question from a different perspective: if there is a hardware fault on any of the quadruped legs, how much of the initial motion's robustness remains?
Next, we explain how we can answer these questions.

Thanks to the way we formulated the \gls{SUF} constraints, we can set individual terms in ${\overline{\bm{K}}_{\bm{\lambda}}}_k$ to zero in order to ``simulate'' what would happen if the torques of the motors of a specific leg of the robot were at their torque limits, or what would happen if the contact force at a specific foot was on the boundary of the friction cone---or both.
We solved \autoref{figure:problems}'s NLP Problem 2 four times (one for each leg) with the goal of analyzing the \gls{SUF} at the end-effector while considering the motor torques to be at their limits and the ground-feet contact forces to be at the boundaries of their respective friction cones.
The dashed lines plotted in \autoref{figure:suf_leg_analysis_bottom} show the results.

We can see that the magnitude of the \gls{SUF} decreases, regardless of which leg is affected.
We can also see that the dashed lines are slightly different from each other, which is expected since the trajectory is not perfectly symmetric and each leg has a slightly different load from every other leg.
Moreover, we can actually identify which leg plays the most important role in this motion by looking at the dashed line with the lowest values (the one for which the \gls{SUF} magnitude decreased the most)---it was the hind right (HR) leg.

\subsection{SUF Optimization with Making and Breaking Contacts}
In this experiment, we wanted to verify that our planning framework is able to optimize trajectories involving switching contacts.
We also evaluated whether the proposed objective function is able to compute a trajectory that is more robust to external disturbances than the baseline cost.

\begin{figure}[ht]
    \captionsetup{font=small}
    \centering
    \includegraphics[width=0.8\linewidth]{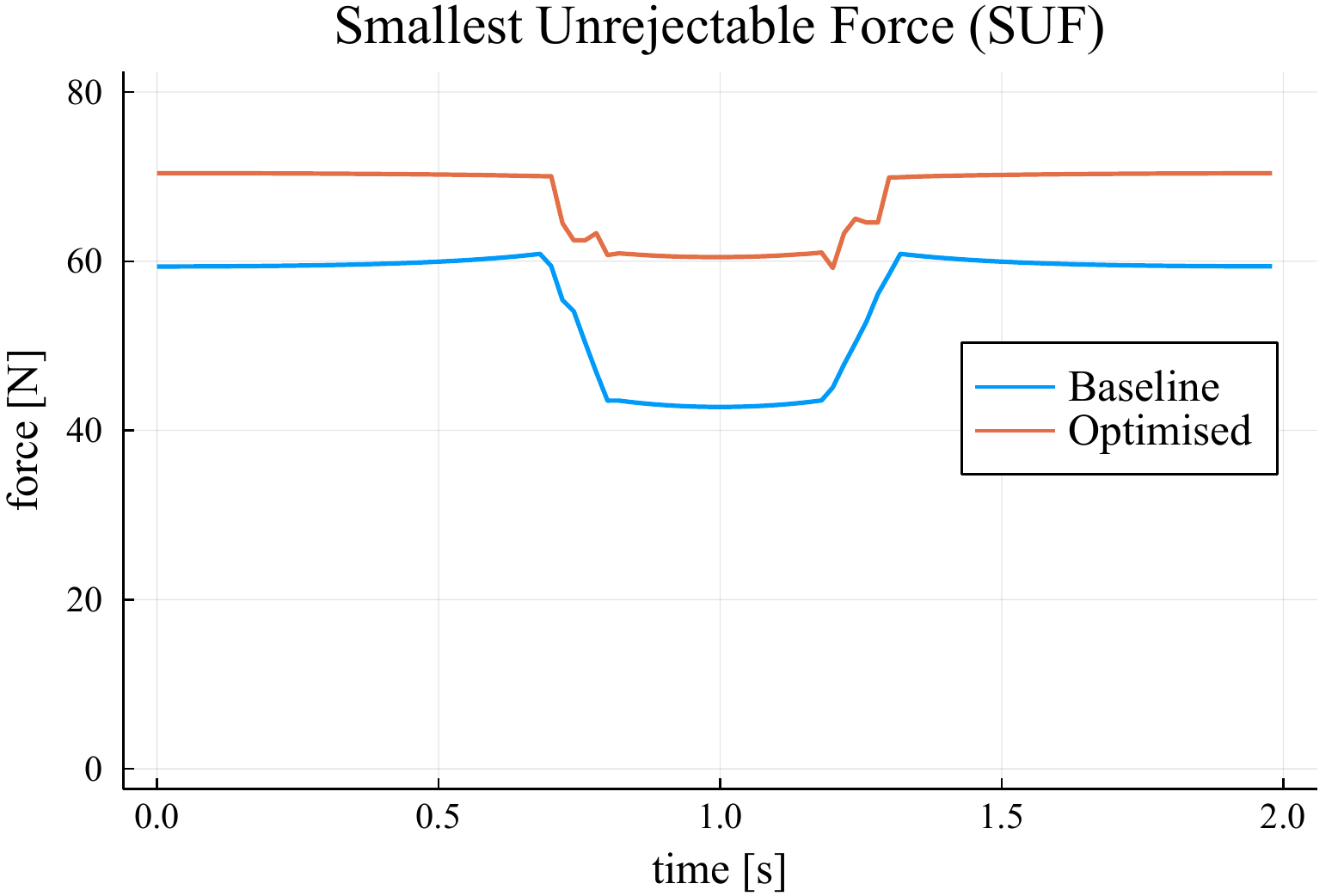}
    \caption{
        Plot of the \gls{SUF} magnitude over time for two trajectories solving the same task specification.
        Blue shows the baseline version (solved with NLP Problem 1), and orange shows the optimised version (solved with NLP Problem 3).
    }\label{figure:suf_step}
\end{figure}

We defined a task in which the robot starts with all four feet on pre-specified positions and then raises its right hind foot off the ground for a short time.
We also constrained the gripper to remain still at a certain location for the duration of the whole motion.
We then solved the task in two different ways:
with NLP Problem 1 (baseline) and with NLP Problem 3 (optimized).
The plot in \autoref{figure:suf_step} shows the magnitude of the \gls{SUF} over time for the resulting trajectories.
Because NLP Problem 1 only outputs a trajectory, we passed it to NLP Problem 2 afterwards in order to compute the \gls{SUF} values---keep in mind that this did not change the baseline trajectory.

The planner was able to compute a feasible trajectory for solving the task using both approaches.
However, as we can see in \autoref{figure:suf_step}, the optimized trajectory is more robust than the baseline.
Moreover, the plot shows that the breaking and making of contacts directly influences the magnitude of the \gls{SUF}.
For both trajectories, the \gls{SUF} magnitude decreases when the right hind foot breaks contact with the floor; and then increases again as the foot reestablishes contact.
Video: \texttt{\small\url{https://youtu.be/H6-g8NLGyYE}}.

\subsection{Robustness Test with Incremental Weights}
In this last (but central) experiment, we plan a robust loco-manipulation trajectory for pulling a rope attached to a bucket and execute it on the real robot.
Our goal is to show that the extended version of our framework can handle problems that require simultaneous locomotion and manipulation while maximizing robustness against disturbances.
Planning loco-manipulation trajectories where robustness is considered proactively was not possible before, and this is really what we have been trying to tackle with our work in this paper.

\begin{figure*}[t]
    \captionsetup{font=small}
    \begin{subfigure}[t]{0.25\linewidth}
        \includegraphics[width=0.99\linewidth]{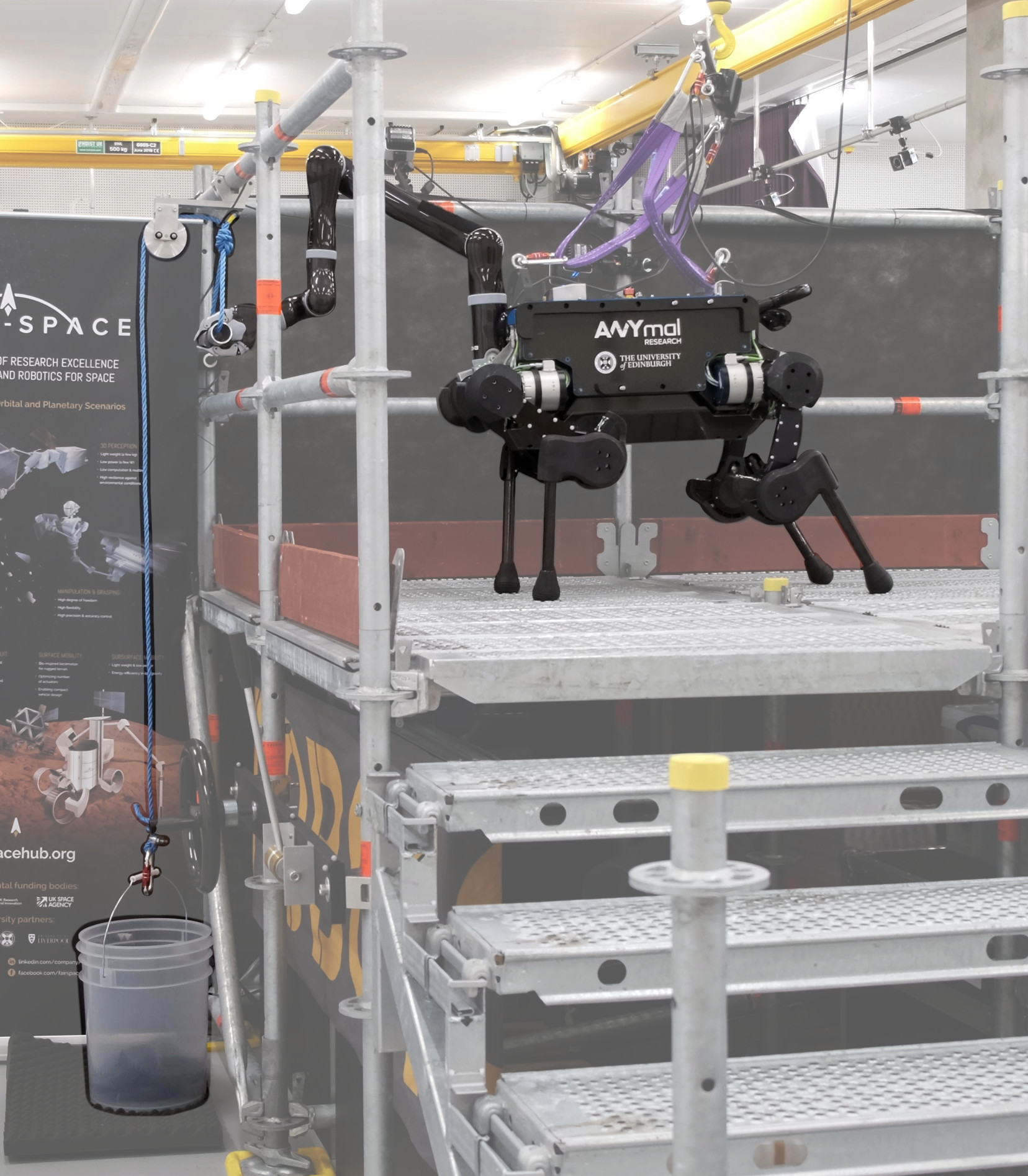}
        \caption{}
    \end{subfigure}\hfill%
    \begin{subfigure}[t]{0.25\linewidth}
        \includegraphics[width=0.99\linewidth]{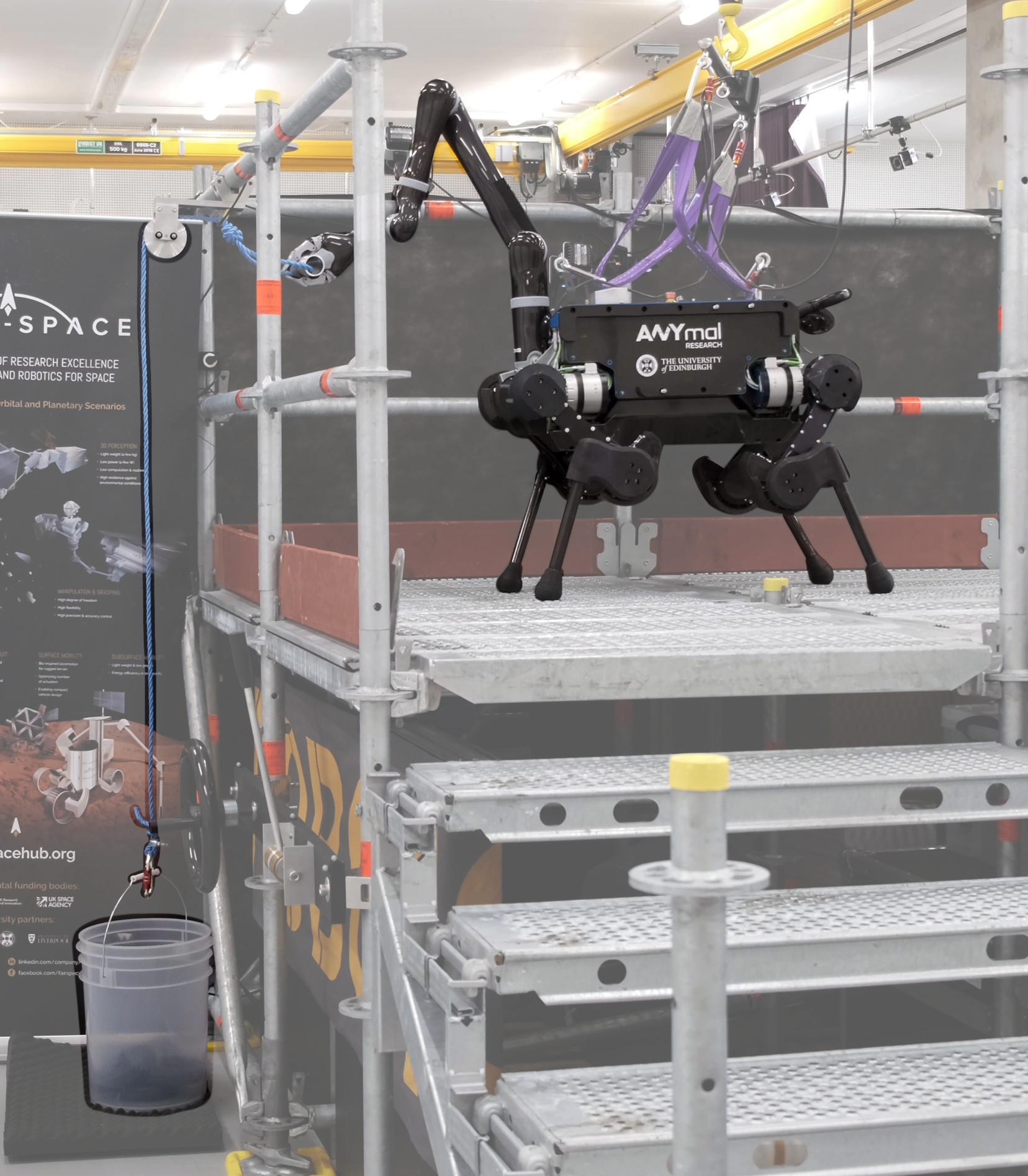}
        \caption{}
    \end{subfigure}\hfill%
    \begin{subfigure}[t]{0.25\linewidth}
        \includegraphics[width=0.99\linewidth]{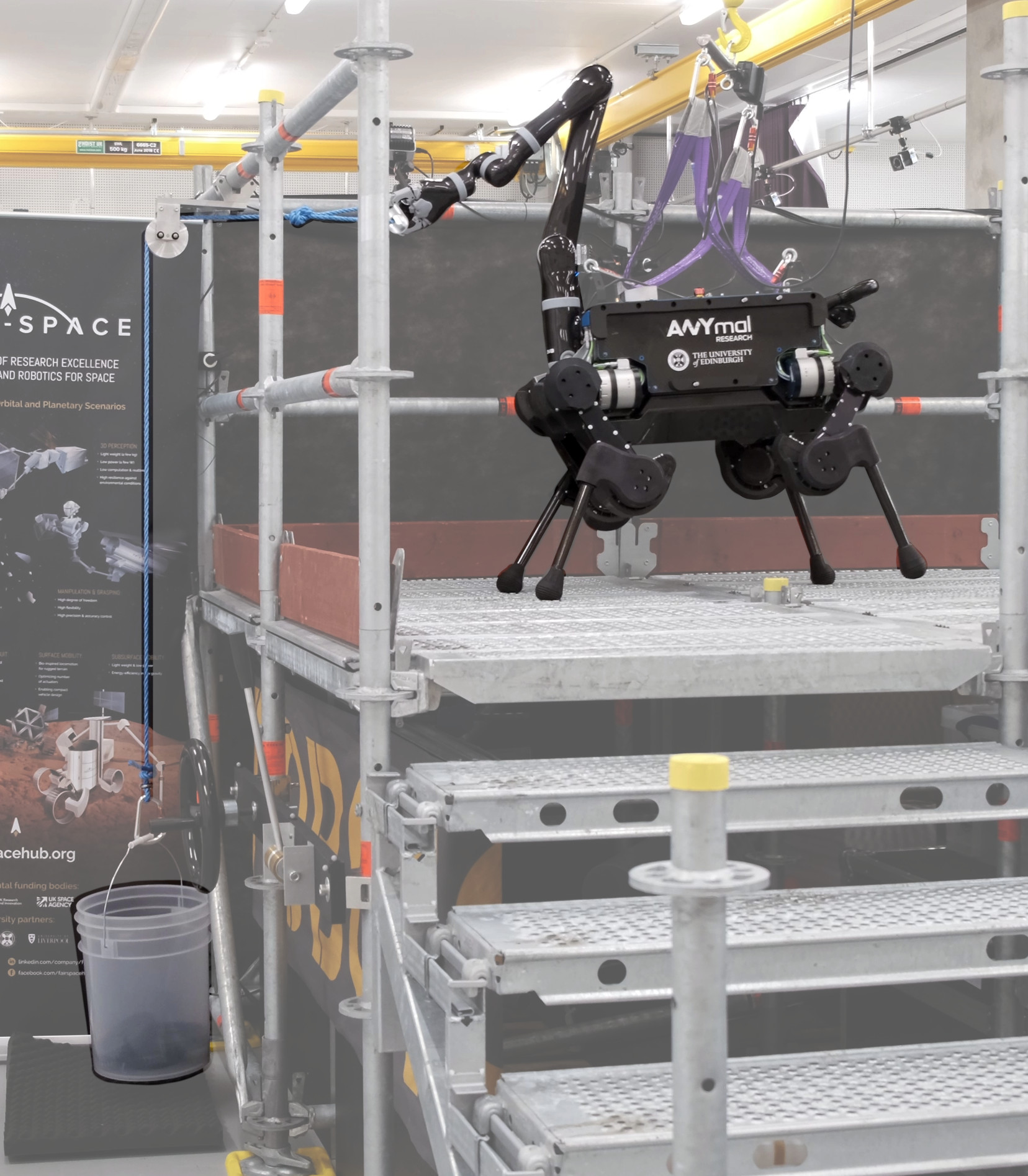}
        \caption{}
    \end{subfigure}\hfill%
    \begin{subfigure}[t]{0.25\linewidth}
        \includegraphics[width=0.99\linewidth]{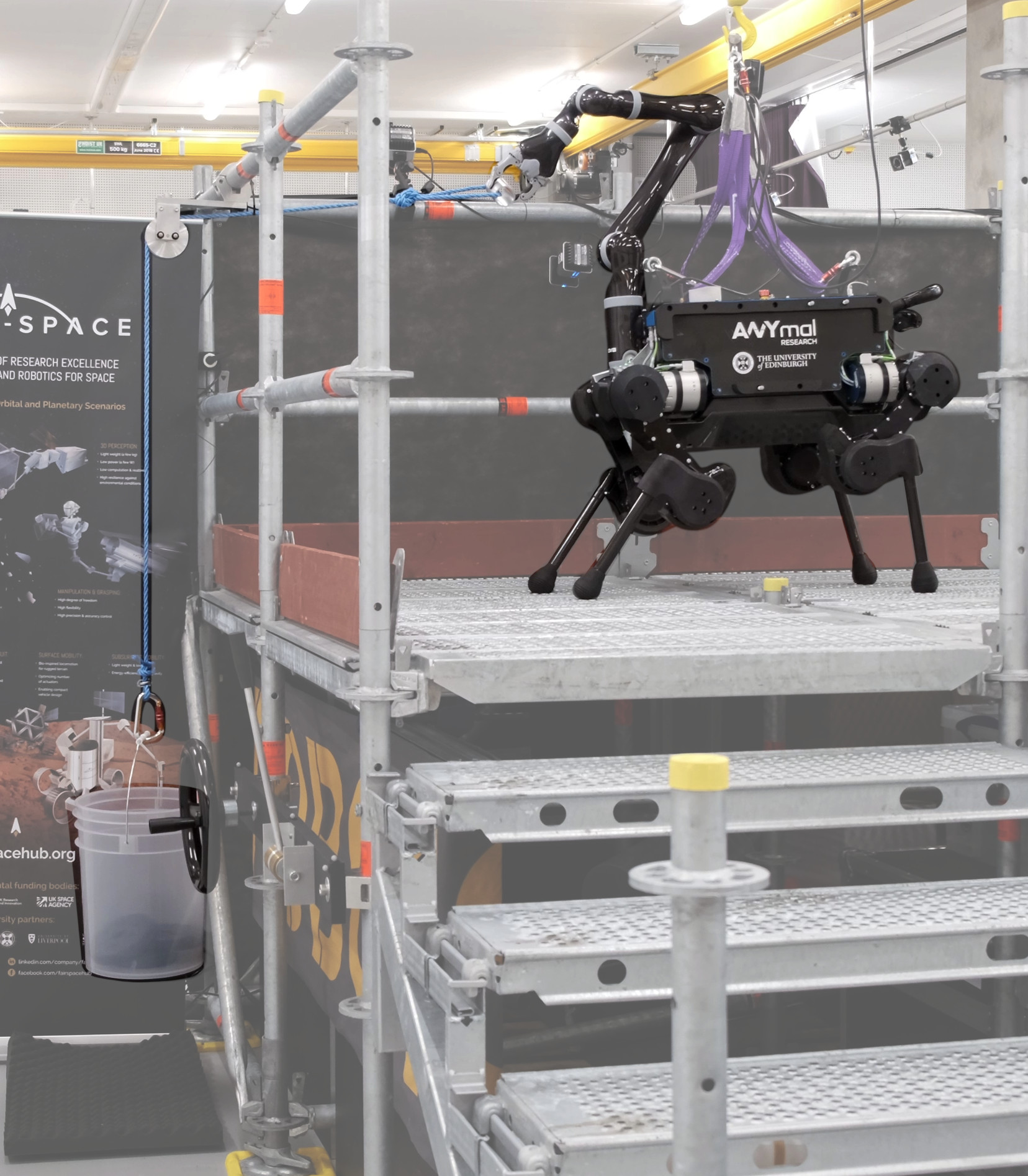}
        \caption{}
    \end{subfigure}
    \caption{
        Snapshots of the robot lifting a bucket with weight plates, weighing \SI{4.1}{\kilo\gram} in total.
        This task requires the quadruped base to take steps at the same time as the arm moves (loco-manipulation).
        Despite the dynamics of the heavy bucket not being modeled, the robot is still able to complete the task, thanks to the robustness of the planned motion and to the feedback gains of the controller.
    }\label{figure:bucket_test}
\end{figure*}

The experimental setup (shown in \autoref{figure:bucket_test}) consisted of a rope threaded through a pulley.
On one end, the rope was attached to a bucket on the ground; and on the other end, the rope was attached to a handle on a platform.
The task for the robot was to grasp the handle from the top of the scaffolding and then pull the rope to lift the bucket.
Importantly, in order to lift the bucket high enough, the robot had to take a few steps back while simultaneously pulling the rope with its arm.
The position of the handle \gls{wrt} the robot was calculated with the Vicon motion capture system (similarly to the other tasks), and the footsteps were pre-specified using a static crawl gait.

\begin{figure}[t]
    \captionsetup{font=small}
    \centering
    \includegraphics[width=0.9\linewidth]{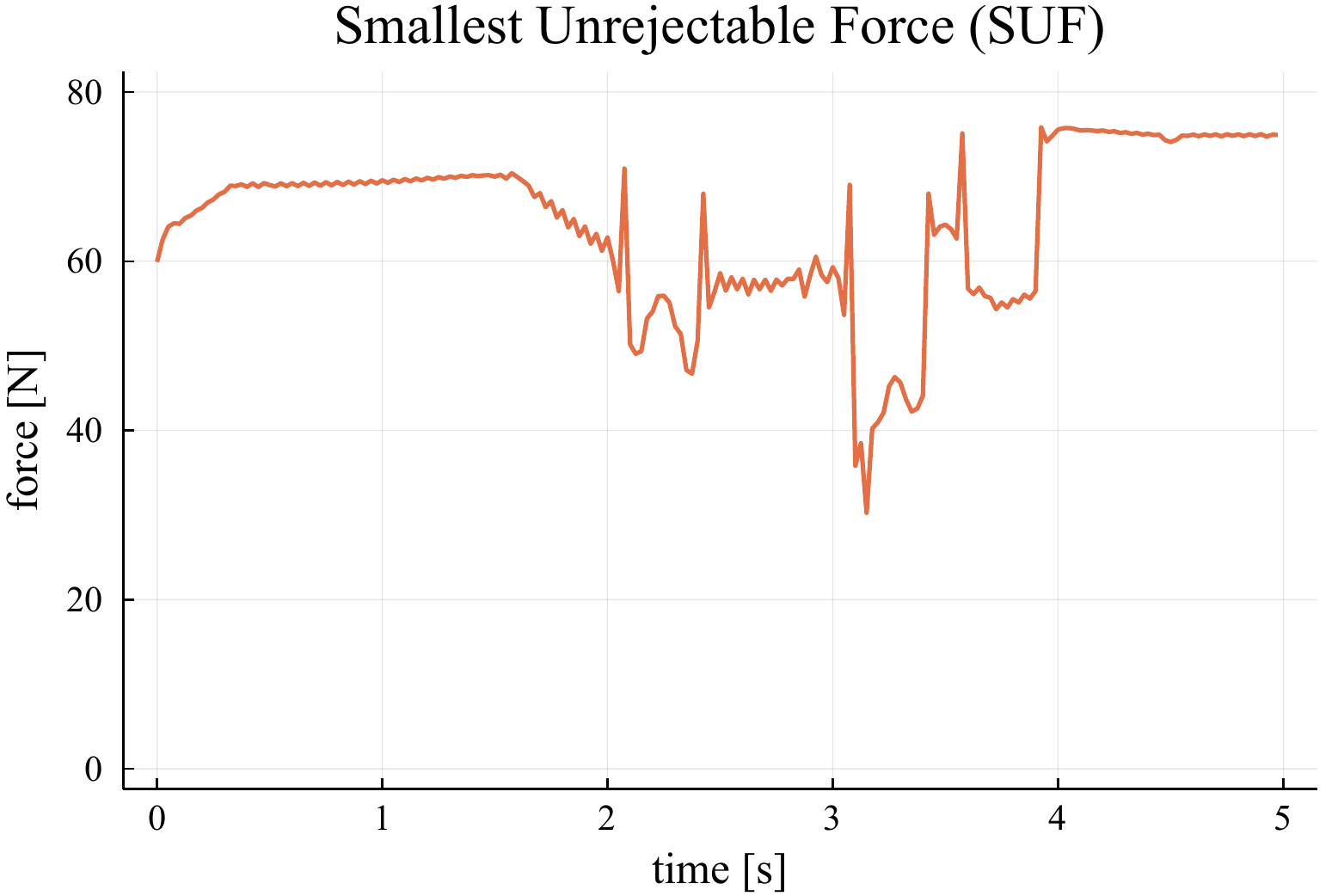}
    \caption{
        \gls{SUF} magnitude over time for the bucket task.
    }\label{figure:bucket_suf}
\end{figure}

To test the robustness of the trajectory planned with our framework, we asked the robot to lift the bucket in multiple trials, wherein the weight of the bucket was incremented \SI{1}{\kilo\gram} in each trial (and the bucket was empty on the first trial).
The bucket itself weighs approximately \SI{1.1}{\kilo\gram}.

The robot completed 4 successful trials, in which it lifted 1.1, 2.1, 3.1, and 4.1 \si{\kilo\gram}.
On the 5th trial (\SI{5.1}{\kilo\gram}) the robot was able to lift the bucket momentarily, but then the handle slipped off the gripper and we counted this as a failure.
A video of the experiment is available: \texttt{\small\url{https://youtu.be/puy2S90_3CM}},
and \autoref{figure:bucket_test} shows some snapshots of the last successful trial (i.e., trial 4), where the robot lifted the bucket containing three \SI{1}{\kilo\gram} weight plates.
Finally, \autoref{figure:bucket_suf} shows the magnitude of the \gls{SUF} over time for this task, where we can see the dips corresponding to the intervals when the robot was taking steps.

This experiment showed that our robot was capable of reliably executing a loco-manipulation trajectory planned with our framework.
Moreover, it showed that the robot is indeed capable of dealing with external disturbances while executing a trajectory.
It is worth remembering that for this experiment (as well as for the other experiments shown in this paper) we did not model the payload; in other words, the system does not know about the bucket or how much it weighs a priori (during planning).
Instead, the weighted bucket is acting as a disturbance, and the robot's ability to lift the bucket is therefore a direct outcome of our robustness metric and the feedback terms used in our controller.

\section{Contact Location Optimization}
\label{sec:contact_location_optimization}

Throughout this work, we pre-specified the position for each foot of the robot during the loco-manipulation planning stage.
For example, when the human operator commands the robot to turn the hand wheel, our framework calculates the feet positions of the robot at that instant via forward kinematics, and then the mathematical constraints of our optimization problem ensure the robot's feet remain on those positions while the wheel is being turned.
But this raises an important question:
when we optimize a trajectory that maximizes robustness, if we constrain the feet to specific positions, will that not limit how robust the resulting trajectory is?
In other words, perhaps the resulting trajectory could be more robust if another set of feet locations had been chosen.

In light of this, we set out to investigate an approach for choosing more adequate contact locations within our motion planning framework.
We now consider the additional problem of optimizing the continuous location of the feet contacts.
Our goal was to understand whether our planner is capable of adapting feet positions such that the resultant whole-body trajectory can be more robust.

\begin{figure*}[t]
    \captionsetup{font=small}
    \begin{subfigure}[t]{0.33\linewidth}
        \includegraphics[width=\linewidth]{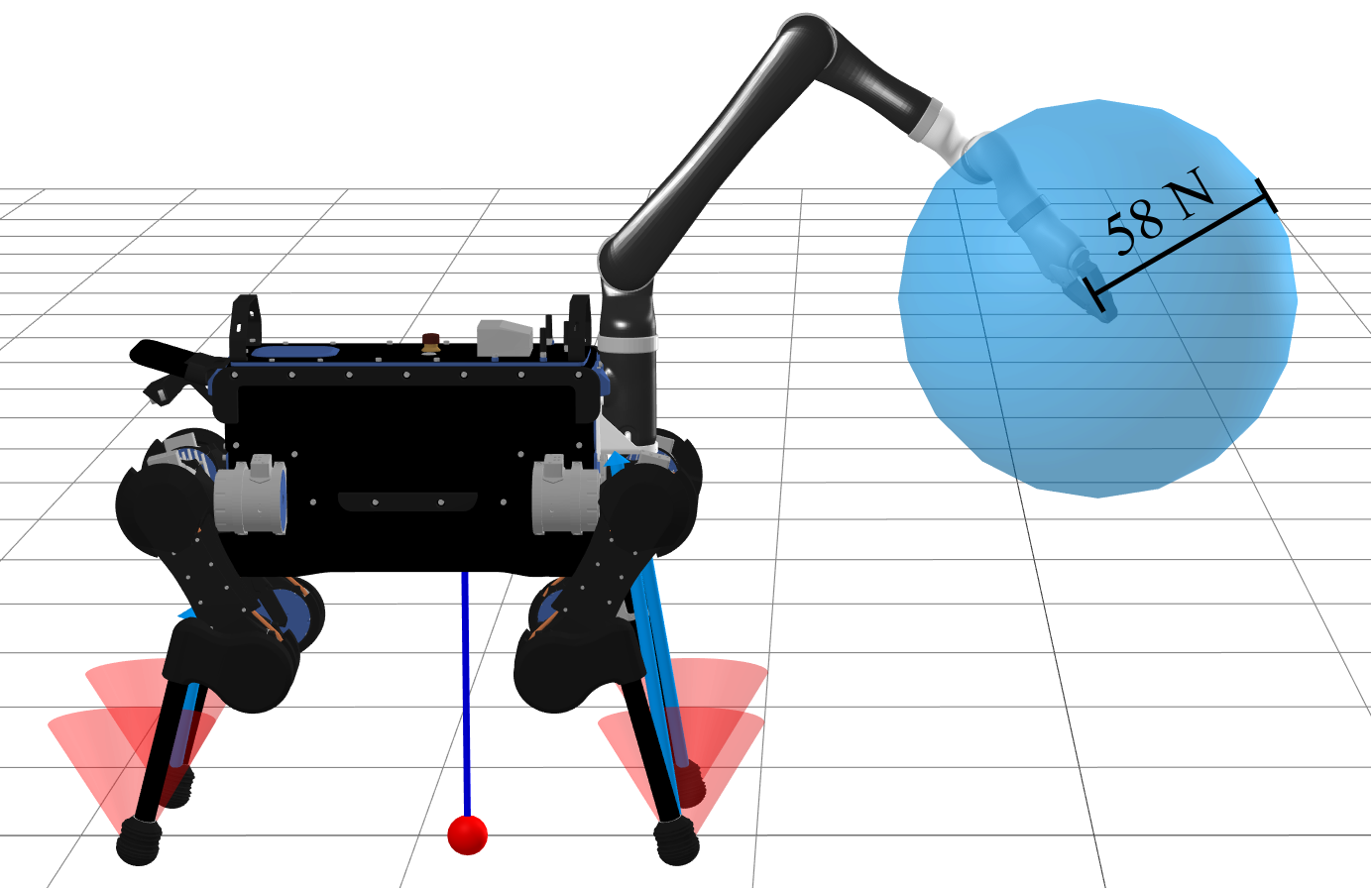}
        \caption{Nominal}
    \end{subfigure}\hfill%
    \begin{subfigure}[t]{0.33\linewidth}
        \includegraphics[width=\linewidth]{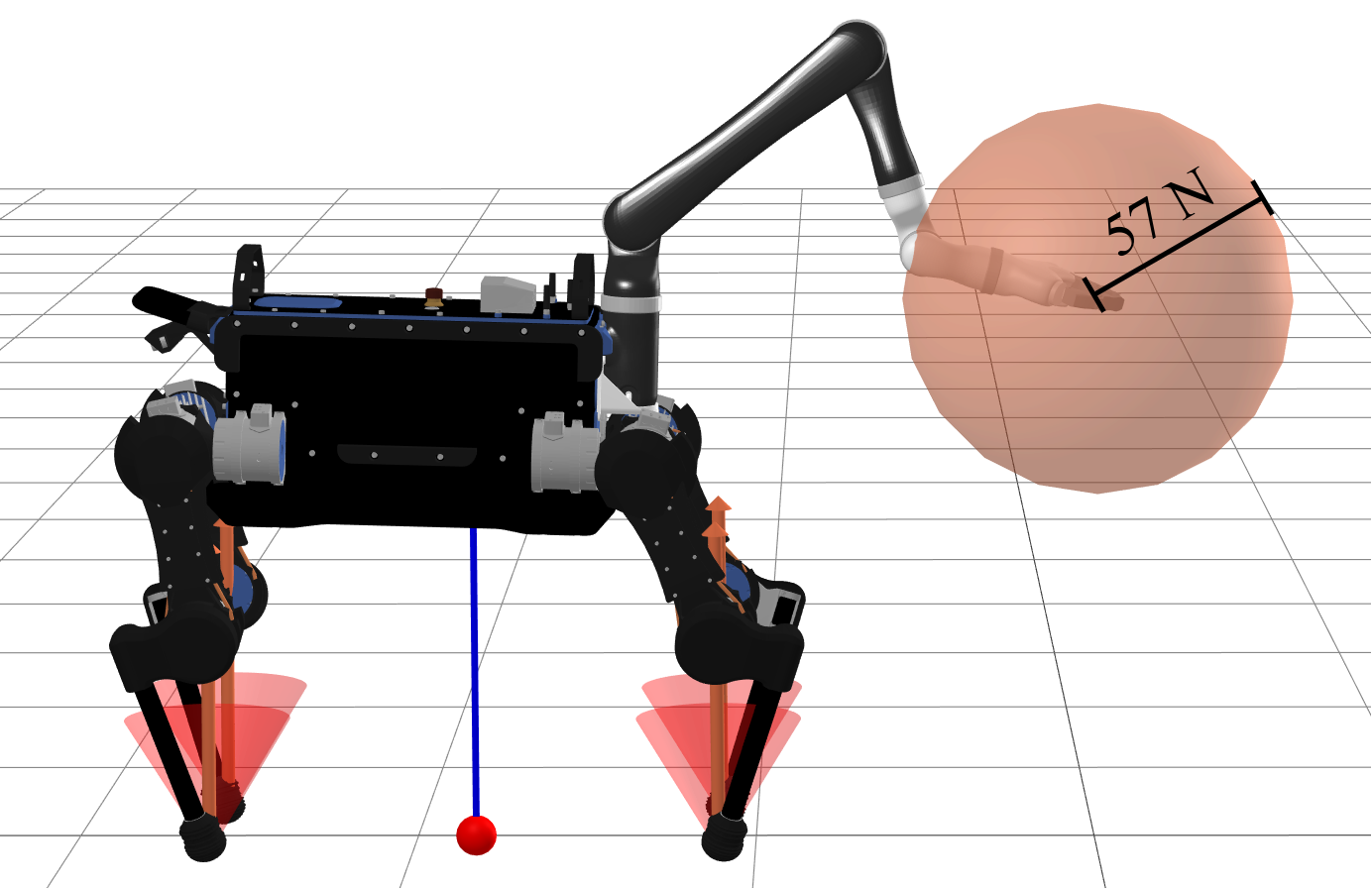}
        \caption{Baseline}
    \end{subfigure}\hfill%
    \begin{subfigure}[t]{0.33\linewidth}
        \includegraphics[width=\linewidth]{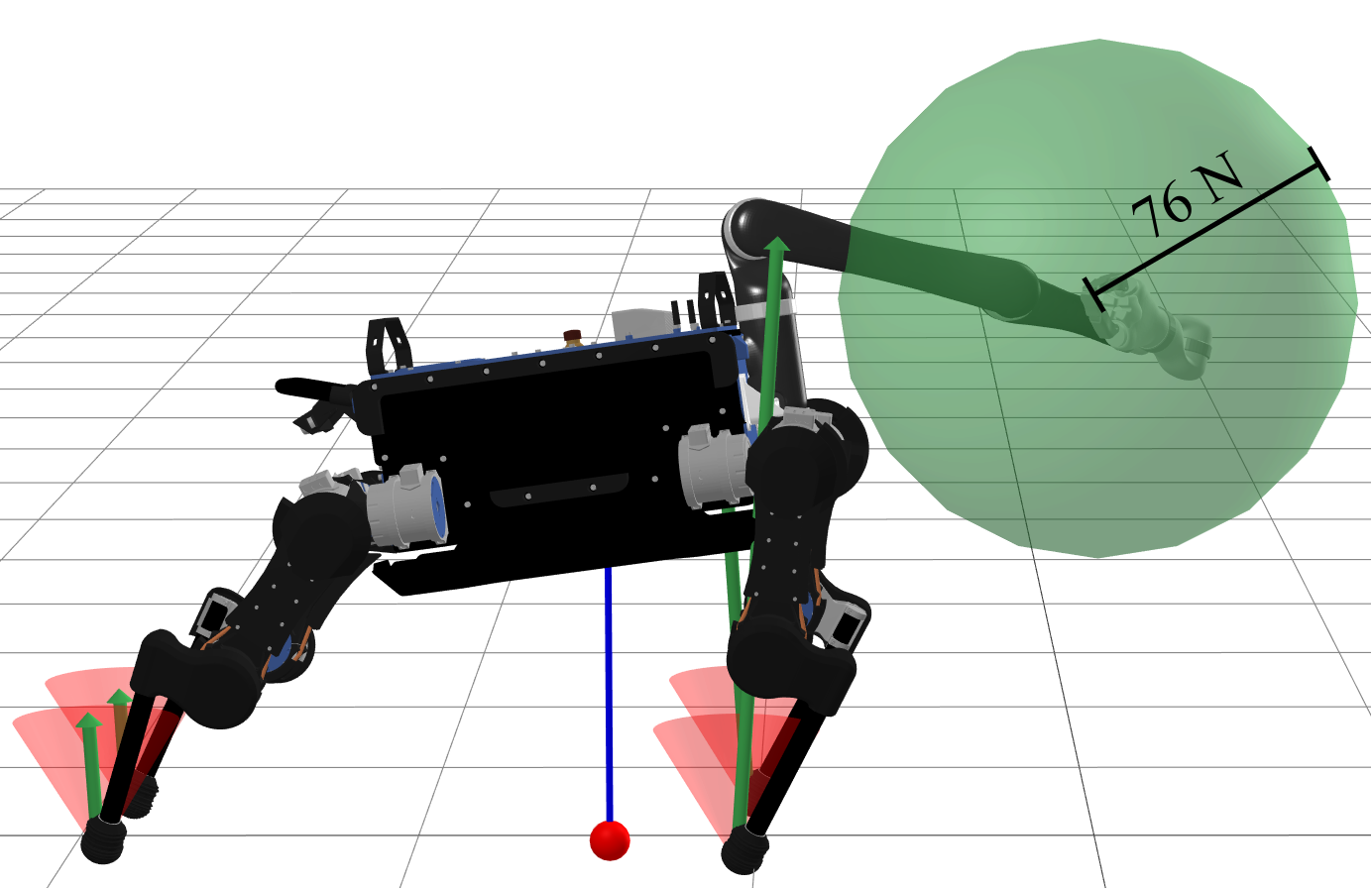}
        \caption{Proposed}
    \end{subfigure}
    \caption{
        Robot configurations resultant from the extended \gls{NLP} formulation which allows the solver to optimize feet locations.
        The radius of the sphere corresponds to the magnitude of the \gls{SUF}.
    }\label{figure:optimised_footsteps_meshcat}
\end{figure*}

In order to enable optimization of feet locations, we extended our \gls{NLP} formulation.
For that, we expand the vector of decision variables of the \gls{NLP} problem to include the $xy$-coordinates of each foot for the beginning of each individual stance phase.
We added the $xy$-coordinates of each foot to our problem as decision variables, assuming that the robot would stand on flat ground (i.e., we assume the $z$-coordinate for each foot is zero).
We also modified \autoref{equation:ee_poses} to consider those decision variables, since previously their right hand-side (i.e., $\bm{p}_i$) were a pre-specified constant position for each foot.
The extended \gls{NLP} formulation is more complex than the one previously used throughout this work: it has additional decision variables (some of them coupled\footnote{Feet positions are represented explicitly by the new $xy$-coordinates, but also implicitly by the forward kinematics of the robot's configuration $\bm{q}$.}), and more complex constraints for enforcing the contact positions; however, it provides more flexibility to the solver, which should now be able to compute feasible trajectories that further maximize robustness by adapting foot contact locations.

To test this more-flexible \gls{NLP} formulation, we defined a motion planning task in which the robot must maintain its configuration while reaching a target point in task-space with its end-effector.
Then, we solved three slightly different versions of the optimization problem:
\begin{itemize}[itemindent=3em]
    \item [\textit{Nominal}] - In this version of the problem, we fixed the configuration of the robot base to its default joint positions (defined by the manufacturer). We did not constrain the configuration of the Kinova arm. Therefore, the solution to this version of the problem consists of the robot moving only its arm to reach for the target and then holding that configuration. No objective function is provided, so this is a feasibility problem.
    \item [\textit{Baseline}] - In this version, the solver is able to change the configuration of the whole-body of the robot. This is similar to the \textit{baseline} formulation used in previous sections, but the feet locations are now decision variables. The torques and contact forces over time are minimized.
    \item [\textit{Proposed}] - In this version, the solver is also able to change the configuration of the whole-body of the robot. This is similar to the \textit{proposed} formulation used in previous sections, but the foot locations are now decision variables. The magnitude of the \gls{SUF} over time is maximized.
\end{itemize}

\begin{figure}[t]
    \captionsetup{font=small}
    \includegraphics[width=\linewidth]{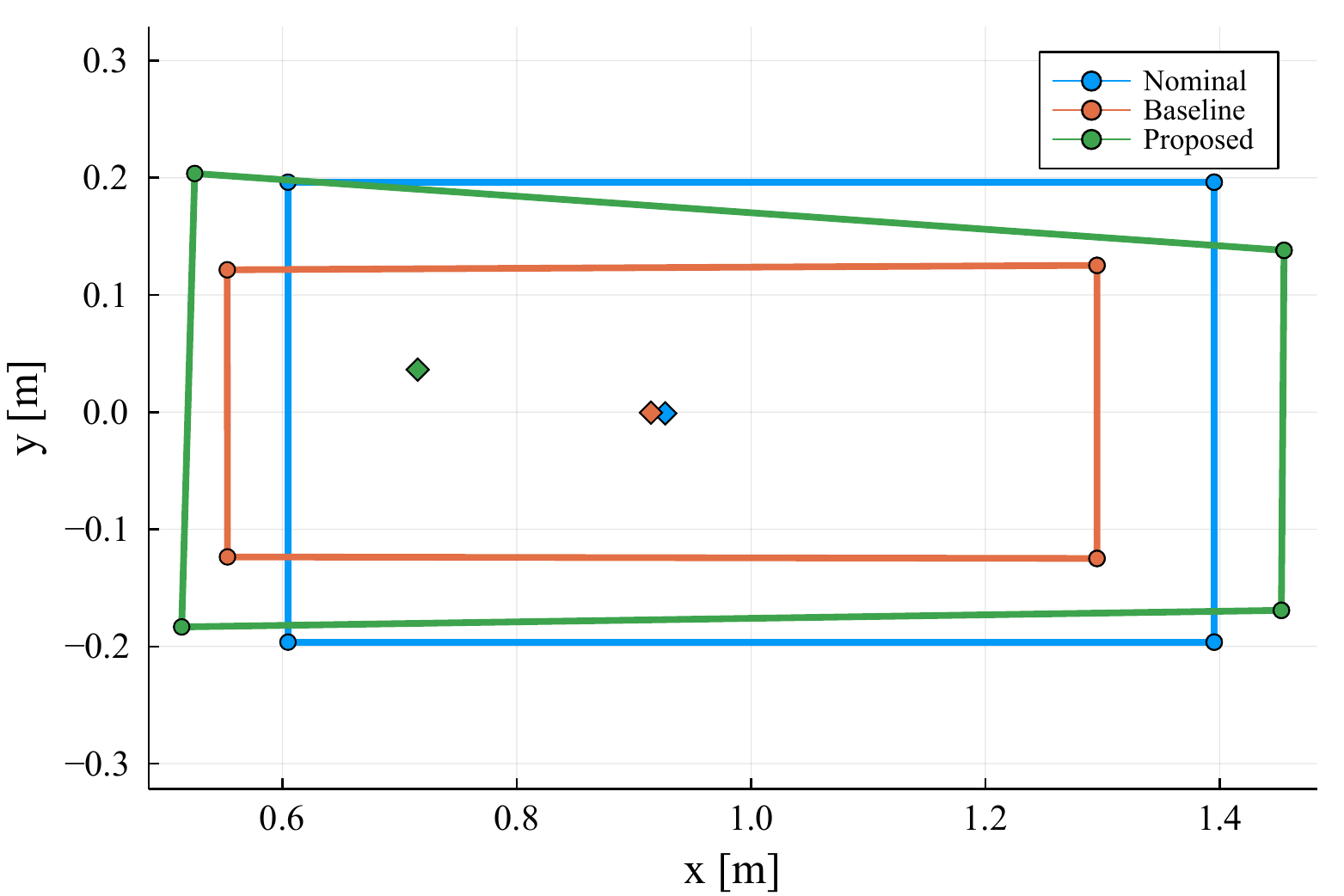}
    \caption{
        Feet locations, support polygons, and projected centers of mass of the trajectories computed using the \textit{nominal} (blue), \textit{baseline} (orange), and \textit{proposed} (green) versions of the optimization problem.
        Feet positions are shown using circles; the line segments connecting the circles form the support polygons; and the diamonds denote the center of mass positions projected onto the support polygons.
    }\label{figure:optimised_footsteps}
\end{figure}

Next, we show the results of these slightly different problem versions.
The robot configurations of each version are shown in \autoref{figure:optimised_footsteps_meshcat}.
The feet positions of each solution are shown in the plot of \autoref{figure:optimised_footsteps}.
After solving each version of the problem, we then computed the magnitude of the \gls{SUF} at the gripper.
The computed \gls{SUF} spheres are also shown in \autoref{figure:optimised_footsteps_meshcat}, and labeled with their respective magnitude.

The first thing we can observe from these results is that the solver did take the liberty of optimizing the feet locations. This verifies that the solver is able to deal with our more-complex \gls{NLP} formulation, albeit taking longer than the previous formulation.
Secondly, the plot of the feet locations clearly shows that the trajectory optimized with the baseline approach has a smaller support polygon than the trajectory optimized with the proposed approach.
This is relevant because we know that the support polygon is a good representation for the region where the \gls{CoM} projection can lie for achieving static balance.
However, the support polygon is only an approximation since it does not take into account robot capabilities (torques and friction at the contacts);
e.g., it is not guaranteed that the robot has enough actuation power to maintain a pose whose \gls{CoM} projection lies on one of the corners of the support polygon.
This leads to our next observation: while the baseline approach keeps the \gls{CoM} position close to the nominal version, the \gls{CoM} of the trajectory optimized with the proposed approach lies further from the center of the support polygon.
Our metric converges to this configuration because it takes into account the robot capabilities, and is able to find a more stable and robust pose, even though its \gls{CoM} projection does not lie at the center of the support polygon.
Finally, we can see that the worst-case disturbance scenario that the robot can resist is better for the proposed approach, and worse for the nominal scenario.
As labeled in \autoref{figure:optimised_footsteps_meshcat}, the magnitude of the \gls{SUF} for the nominal, baseline, and proposed approaches were \SI{58}{\newton}, \SI{57}{\newton}, and \SI{76}{\newton}, respectively.

The above results show that our formulation is able to optimize feet locations, and that our metric is able to guide the solver to solutions that have increased capabilities of resisting external forces---at least in theory, that is.
However, when we deploy our method on the robot, there are always practical subtleties that may affect how the robot behaves, such as signal delay or the type of controller being used.
With that in mind, we carried out an experiment to assess the actual force-rejection capabilities of the real robot.
Next, we explain the experiment setup and then analyze the results.

\begin{figure*}[t]
    \captionsetup{font=small}
    \includegraphics[width=\linewidth]{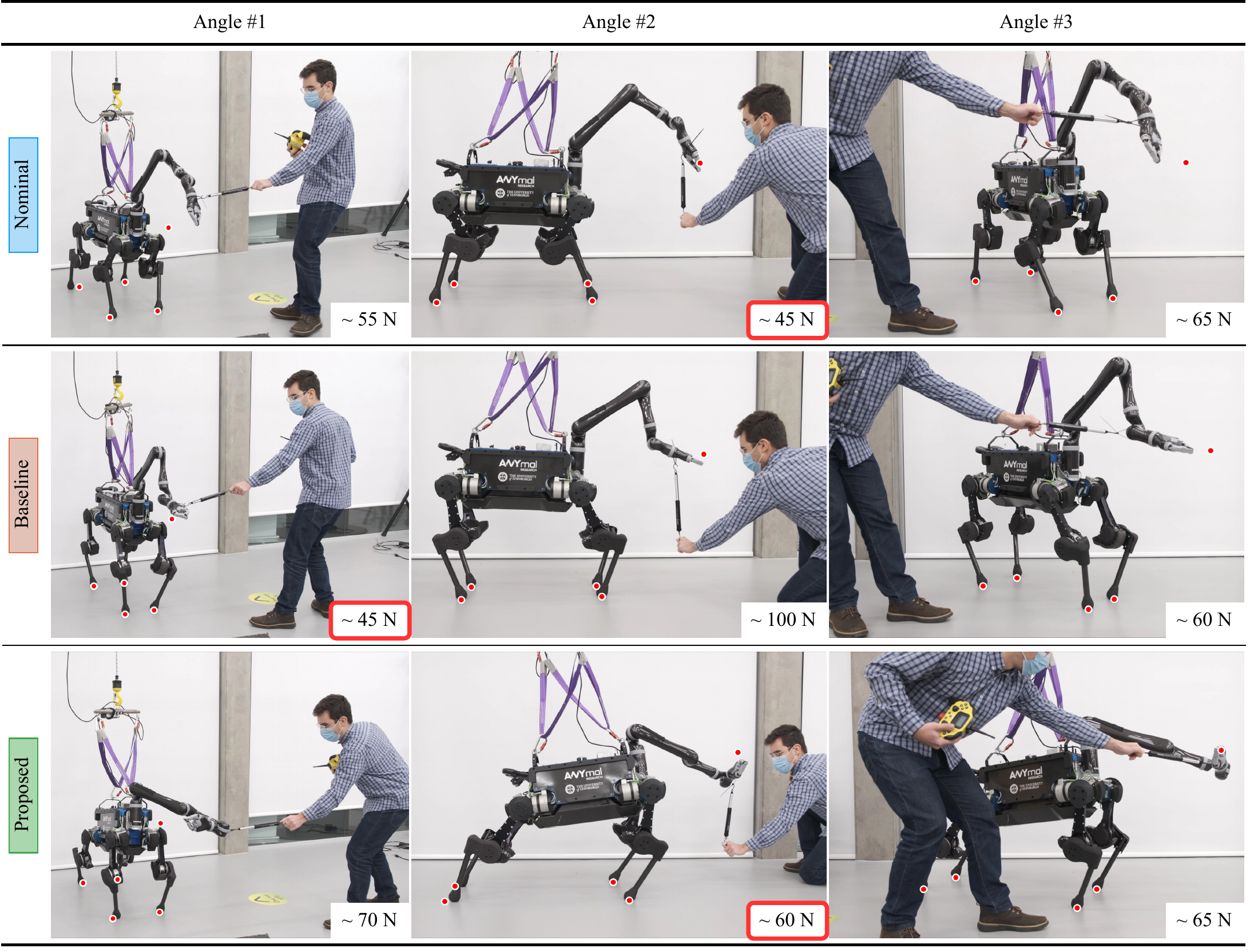}
    \caption{
        Snapshots of the experiment in which we disturbed the robot by pulling its end-effector.
        Pictures on the same row correspond to the same trajectory, whereas pictures under the same column share the same disturbance angle.
        The snapshots were taken from the video footage collected during the experiment, at the time the force gauge indicated the greatest force magnitude---which is shown in the label on the bottom right corner of each picture.
        Labels highlighted in red are the worst-case scenario for each row (out of the three angles shown).
        The five red circles on each snapshot represent the initial positions of the four feet and of the end-effector.
        Video: \texttt{\small\url{https://youtu.be/tUXQUqLneTE}}.
    }\label{figure:optimised_footsteps_snapshots}
\end{figure*}

In this experiment, we executed the trajectory optimized with each approach on the real robot and, for each case, we disturbed the robot by pulling its end-effector from three different angles.
A summary of the events that occurred for each angle/trajectory is given below:

\begin{itemize}[itemindent=-1em]
    \item [] \textit{Summary of Angle \#1}
          \begin{itemize}[itemindent=3em]
              \item [\textit{Nominal}] - First, the end-effector deviated from its set point. Then, the right hind and right front feet started to slide. However, the robot did not fall over.
              \item [\textit{Baseline}] - First, the end-effector deviated from its set point. Then, the right front foot started to slide. Next, the right hind foot lost contact. Finally, the robot toppled to its left side.
              \item [\textit{Proposed}] - First, the end-effector deviated from its set point. Then, the right front foot started to slide. Next, both left and right hind feet slipped, but only slightly. The robot did not fall over.
          \end{itemize}
    \item [] \textit{Summary of Angle \#2}
          \begin{itemize}[itemindent=3em]
              \item [\textit{Nominal}] - The end-effector deviated from its set point and the hind feet started to slip at approximately the same time. The robot would have fallen down, but the harness prevented it from collapsing on the floor.
              \item [\textit{Baseline}] - The end-effector deviated from its set point. The feet did not move and the robot did not fall over.
              \item [\textit{Proposed}] - First, the end-effector deviated from its set point. Then, the hind feet started to move. If the harness had not supported it, the robot would have fallen down.
          \end{itemize}
    \item [] \textit{Summary of Angle \#3}
          \begin{itemize}[itemindent=3em]
              \item [\textit{Nominal}] - The end-effector deviated from its set point. The left hind foot moved. The robot did not fall.
              \item [\textit{Baseline}] - The end-effector deviated from its set point. The left front foot moved and the left hind foot lost contact. Finally, the robot toppled over to its right side.
              \item [\textit{Proposed}] - The end-effector deviated from its set point. The left front foot moved and the left hind foot lost contact. Finally, the robot toppled over to its right side.
          \end{itemize}
\end{itemize}

When we carried out the experiment, we used a force gauge between the end-effector of the robot and the source of the disturbance in order to measure---and capture on video---the magnitude of the force being applied to the robot's end-effector.

In \autoref{figure:optimised_footsteps_snapshots}, we show the state of the experiment at the instant when the force gauge indicated the greatest force magnitude, for each trajectory and for each angle.
The label on the bottom right corner of each snapshot indicates the magnitude of the value measured by the force gauge.
For each row, i.e., for each trajectory, the label highlighted in red corresponds to the magnitude of the \gls{SUF} found experimentally for the three distinct angles used for the experiment.
The way to interpret these results is to take the smallest \gls{SUF} found experimentally for each trajectory (highlighted in red and signifying the worst case perturbation), and then compare those magnitudes to find the trajectory with largest worst case \gls{SUF} (indicating the highest robustness).

As highlighted in \autoref{figure:optimised_footsteps_snapshots}, the proposed trajectory exhibited the highest \gls{SUF} (\SI{60}{\newton}).
In percentage terms, the \gls{SUF} of the proposed trajectory was approximately $\SI{33}{\percent}$ better than both nominal and baseline trajectories, which is a significant improvement.
Moreover, the \gls{SUF} magnitudes found experimentally were \qtyrange{21}{22}{\percent} smaller than the \gls{SUF} predicted originally (the ones shown in \autoref{figure:optimised_footsteps_meshcat}).
We understand this is due to model mismatch and unaccounted factors when we deploy the trajectory on the real robot.
But the fact that the percentage decrease is similar for each trajectory is reassuring, as it tells us that the unaccounted factors of the real robot affected the \gls{SUF} equally, and the proposed trajectory performed better than the nominal and baseline trajectories as we had predicted in relative terms.

\section{Discussion and Future Work}
In this paper, we presented a framework for planning whole-body loco-manipulation trajectories robust to external disturbances.
We integrated our framework with existing software stacks to enable easy switching between teleoperated and autonomous modes.
We demonstrated the capabilities of that integration by having a human operator remotely control the robot via a joystick in a mock-up rig of an industrial site, and by having the robot autonomously plan complex and rich whole-body motions for real-world tasks within that setting, such as turning a hand wheel, pulling a lever, opening a gate whilst on a ramp, and lifting a heavy bucket by pulling a rope.
We also carried out a wide range of experiments to test the reliability of the full system, analyze the \gls{SUF} of existing trajectories, optimize the robustness of trajectories (including those involving making and breaking of contacts).
Finally, we carried out an initial investigation on the possibility of using our framework for the purpose of optimizing feet positions.

Our experiments showed that the resulting system was reliable and versatile.
The robot was able to navigate the industrial scaffolding and successfully execute all the challenges we prepared.
Moreover, we showed that the \gls{SUF} is a valuable metric, both for analyzing existing trajectories and for maximizing the robustness of new trajectories.
Finally, our preliminary research on the optimization of feet locations showed promising results.
We now conclude with a list of interesting avenues for future work.

\subsection{Taking Into Account Force-Feedback During Execution}
In this work, we have disregarded the dynamics model of objects being manipulated by the robot.
We maximized robustness against external disturbances at the end-effector in order to demonstrate that we do not have to necessarily model the object dynamics (which may not be available at planning time), as the controller is able to track the reference trajectory and compensate for the object dynamics through feedback terms at the controller level.
However, in extreme cases, controller feedback terms are not enough to execute the desired motion appropriately.

An immediate extension of our work would be to take into account force-feedback during execution of the task.
We could plan the motion just like we have done throughout this paper before starting the actual manipulation task.
However, once the robot starts executing the task, the object being manipulated will exert a set of forces that in turn apply torque at the robot's joints.
An interesting direction for future work would be to take into account those torques at the joint level to estimate the force being applied to the robot as a consequence of the robot-object interaction, and then replanning the motion taking into account that force estimation.
This would mean that we would still be able to maximize the \gls{SUF}, but we would now have a much better model of the task taking place.

\subsection{Decreasing Computational Time for Real-Time Execution}
For many real-world tasks, such as the ones we showed with our robot, it is sufficient to take the current state of the world, plan a motion in ``one-shot'' for completing the task, and then executing that trajectory.
However, there are scenarios where this approach might fail, especially if the state of the world changes as the robot executes the task---meaning that the planned trajectories become invalid.
Real-time control schemes (such as MPC) are usually employed for dealing with those scenarios.
Currently, the computation times of our framework for planning robust trajectories are too slow to be compatible with the budget available in those real-time control schemes.

There are a few possible approaches for tackling this challenge, if we want to use our framework for such scenarios.
In the first approach, we can try to decrease the computational time e.g. by making the robot model simpler, making the problem smaller by considering a more coarse discretization, or by decreasing the complexity of the problem constraints.
Another approach would be to change the NLP solver used for solving the trajectory optimization problems. E.g., we could attempt to formulate our robustness optimization problem using a DDP-based approach, such as \cite{mastalli2020crocoddyl}, instead of using a direct transcription approach that relies on off-the-shelf solvers. However, the disadvantage of this approach is that, while off-the-shelf commercial solvers are able to deal with a wide range of constraints, enforcing general constrains with DDP-based approaches is pretty much an active research topic of its own.
Finally, a third possible approach would be to speed-up the computation of the robustness metric itself. Perhaps there is a way the metric can be learnt, represented by a surrogate model, or stored in a look-up table for quick retrievals.

\subsection{Whole-Body Robustness Analysis}
In this work and in our previous work \cite{ferrolho2020optimizing}, we focused on improving the robustness to external disturbances applied at the end-effector.
This was because we were interested in applications where the robot had to interact with objects without modeling the dynamics of those objects.
However, the mathematical derivation of the \gls{SUF} applies to any point on any rigid body of the robot mechanism.
In other words, instead of computing the \gls{SUF} at the end-effector, we could also compute the \gls{SUF} at the robot's base, at the knees, at the elbows, and so on.
An interesting path for future work would be the development of a software for analyzing robot trajectories where a robot and a trajectory are given as inputs, and then as an output we could see the \gls{SUF} spheres at multiple points along the robot's kinematic chain.
This kind of visualization would allow us to better understand and have a very clear visual representation of bottlenecks and weaknesses in trajectories.
With such a tool, we would be able to look at a trajectory, and easily indicate that at a certain time, the robot is susceptible to failure even with small disturbances applied to specific parts of its body.

\subsection{Robust Footstep Optimization}
Another interesting direction for future research is the possibility of finding contact locations by taking into account the robustness metric.
We presented preliminary work on this topic in the previous section.

\appendix

In \autoref{table:supplementary_videos}, we list all the videos supplementing our work, in the same order as they appear in this manuscript.
A playlist is also available here: \texttt{\small\href{https://shorturl.at/oFJU0}{shorturl.at/oFJU0}}.

\begin{table}[ht]
    \captionsetup{font=small}
    \centering
    \caption{Supplementary videos.}
    \label{table:supplementary_videos}
    \begin{tabular}{ll}
        \toprule
        Description                   & Link (YouTube)                                                      \\
        \midrule
        System demonstration          & \texttt{\href{https://youtu.be/3qXNHVCagL8}{youtu.be/3qXNHVCagL8}}  \\
        Repeatability experiment      & \texttt{\href{https://youtu.be/Ok8Pcwn_I0w}{youtu.be/Ok8Pcwn\_I0w}} \\
        Robustly turning a wheel      & \texttt{\href{https://youtu.be/1M32AHuuDhI}{youtu.be/1M32AHuuDhI}}  \\
        Robustly pulling a lever      & \texttt{\href{https://youtu.be/6A9eSdfcj7A}{youtu.be/6A9eSdfcj7A}}  \\
        SUF with contact switching    & \texttt{\href{https://youtu.be/H6-g8NLGyYE}{youtu.be/H6-g8NLGyYE}}  \\
        Test with incremental weights & \texttt{\href{https://youtu.be/puy2S90_3CM}{youtu.be/puy2S90\_3CM}} \\
        Robust footstep locations     & \texttt{\href{https://youtu.be/tUXQUqLneTE}{youtu.be/tUXQUqLneTE}}  \\
        \bottomrule
    \end{tabular}
\end{table}

\bigskip\textbf{Acknowledgments} \
We would like to thank Douglas Howie and Laura Ferguson for their hard work in fabricating the props for the experiments shown in this paper.
We would also like to thank Xinnuo Xu and Traiko Dinev for their help recording the videos of our experiments.
Finally, we thank the Reviewers for their constructive comments on this manuscript.

\bigskip\textbf{Author Contributions} \
Henrique Ferrolho led the research, proposed the core concepts, implemented the framework, and wrote the manuscript.
Vladimir Ivan helped with the research, engaged actively in technical discussions, and helped writing the manuscript.
Wolfgang Merkt also took part in technical discussions, developed the controller used in our experiments, and helped writing the manuscript.
Sethu Vijayakumar and Ioannis Havoutis provided critical feedback and helped shape the research, analysis, and manuscript.

\bigskip\textbf{Funding} \
This research was supported by:
The Alan Turing Institute,
EPSRC UK RAI Hub for Offshore Robotics for Certification of Assets (ORCA, EP/R026173/1),
EU H2020 project Memory of Motion (MEMMO, 780684),
EPSRC as part of the Centre for Doctoral Training in Robotics and Autonomous Systems at Heriot-Watt University and The University of Edinburgh (EP/L016834/1).
This research has received funding from the EU H2020 research and innovation programme under grant agreement No 101017008, Enhancing Healthcare with Assistive Robotic Mobile Manipulation (HARMONY).

\bigskip\textbf{Conflict of Interest} \
The authors have no competing interests to declare that are relevant to the content of this article.

\ifCLASSOPTIONcaptionsoff
    \newpage
\fi

\bibliographystyle{IEEEtran}
\bibliography{IEEEabrv,references}


\begin{IEEEbiography}[{\includegraphics[width=1in,height=1.25in,clip,keepaspectratio]{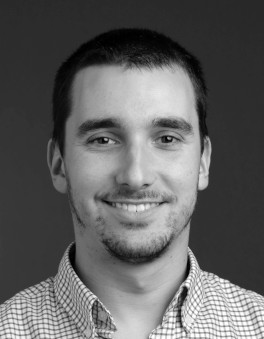}}]
    {Henrique Ferrolho} received his Ph.D. in robotics and autonomous systems from The University of Edinburgh, Edinburgh, U.K. (2021) and B.Sc./M.Sc. in informatics and computing engineering from the University of Porto, Porto, Portugal (2017).
    He is currently a Robotics Engineer at Ocado Technology, developing robot manipulation solutions for picking and placing tens of thousands of grocery products of varying shapes, sizes, weights, and fragility.
\end{IEEEbiography}

\begin{IEEEbiography}[{\includegraphics[width=1in,height=1.25in,clip,keepaspectratio]{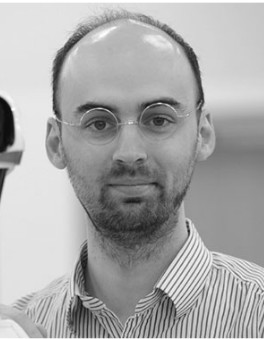}}]
    {Vladimir Ivan} received the B.Sc. degree in artificial intelligence (AI) and robotics from the University of Bedfordshire, Luton, U.K., in 2009, the M.Sc. degree in AI specializing in intelligent robotics, in 2010 and the Ph.D. degree in motion synthesis in topology-based representations from The University of Edinburgh, Edinburgh, U.K., in 2014 respectively, where he then worked as a senior researcher until 2022. His research interests include motion planning and modelling, topology, humanoid and legged robotics, space robotics, and shared autonomy.
    He is currently VP of Robotics at Touchlab Limited in Edinburgh, developing tactile tele-robots.
\end{IEEEbiography}

\begin{IEEEbiography}[{\includegraphics[width=1in,height=1.25in,clip,keepaspectratio]{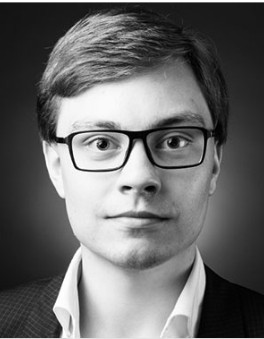}}]
    {Wolfgang Merkt} received the B.Eng. (Hons.) degree in mechanical engineering with management and the M.Sc. (R) and Ph.D. degrees in robotics and autonomous systems from The University of Edinburgh, Edinburgh, U.K., in 2014, 2015 and 2019, respectively.
    Subsequently, he has been with the Oxford Robotics Institute, University of Oxford, Oxford, U.K. from 2020--2023, most recently as a Senior Researcher.
    During his Ph.D., he worked on trajectory optimization and warm-starting optimal control for high-dimensional systems and humanoid robots.
    His research interests include optimization- and learning-based methods for planning and control, loco-manipulation, and legged robots.
\end{IEEEbiography}

\begin{IEEEbiography}[{\includegraphics[width=1in,height=1.25in,clip,keepaspectratio]{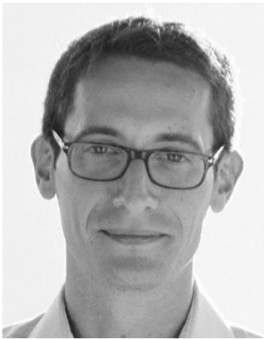}}]
    {Ioannis Havoutis} received the Ph.D. in Informatics (2011) and M.Sc. in Artificial Intelligence (2007) from The University of Edinburgh.
    He is currently a Lecturer in Robotics with the University of Oxford, Oxford, U.K.. He is part of the Oxford Robotics Institute and a colead of the Dynamic Robot Systems Group.
    His focus is on approaches for dynamic whole-body motion planning and control for legged robots in challenging domains.
    From 2015 to 2017, he was a Postdoc with the Robot Learning and Interaction Group, Idiap Research Institute, Martigny, Switzerland.
    From 2011 to 2015, he was a Senior Postdoc with the Dynamic Legged System Laboratory, Istituto Italiano di Tecnologia, Genoa, Italy.
\end{IEEEbiography}

\begin{IEEEbiography}[{\includegraphics[width=1in,height=1.25in,clip,keepaspectratio]{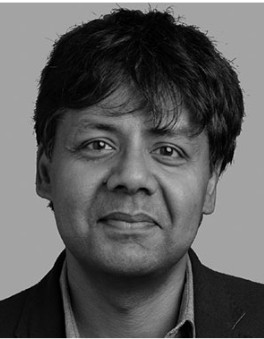}}]
    {Sethu Vijayakumar} received the Ph.D. degree in computer science and engineering from the Tokyo Institute of Technology, Tokyo, Japan, in 1998.
    He is Professor of Robotics and Founding Director of the Edinburgh Centre for Robotics, where he holds the Royal Academy of Engineering Microsoft Research Chair in Learning Robotics within the School of Informatics at the University of Edinburgh, U.K.
    He also has additional appointments as an Adjunct Faculty with the University of Southern California, Los Angeles, CA, USA and a Visiting Research Scientist with the RIKEN Brain Science Institute, Tokyo.
    His research interests include statistical machine learning, whole body motion planning and optimal control in robotics, optimization in autonomous systems as well as optimality in human motor motor control and prosthetics and exoskeletons.
    Professor Vijayakumar is a Fellow of the Royal Society of Edinburgh.
    In his recent role as the Programme Director for Artificial Intelligence and Robotics at The Alan Turing Institute, Sethu helps shape and drive the UK national agenda in Robotics and Autonomous Systems.
\end{IEEEbiography}




\end{document}